\documentclass[5p,times]{elsarticle}

\usepackage{lineno,hyperref}
\usepackage{amssymb}
\usepackage{array}
\usepackage{caption}
\usepackage{subcaption}
\usepackage{xcolor}
\usepackage{booktabs}
\usepackage{multirow}
\usepackage{float}
\usepackage{siunitx}
\usepackage{textcomp}
\usepackage{makecell}
\usepackage{listings}
\usepackage[singlelinecheck=false ]{caption}
\usepackage{textcomp }
\usepackage{multirow}
\usepackage{hyperref}
\usepackage[normalem]{ulem}
\usepackage{amsmath}
\usepackage{xurl}
\usepackage{algorithm}
\usepackage[noend]{algpseudocode}
\usepackage{graphicx}
\usepackage{tabularx}
\usepackage{manfnt}
\usepackage{apalike}
\usepackage{geometry}
\useunder{\uline}{\ul}{}

\DeclareMathOperator*{\argmaxA}{arg\,max} % Jan Hlavacek
\modulolinenumbers[5]
\DeclareMathOperator*{\argmin}{argmin}

\newtheorem{definition}{Definition}
\begin{document} \sloppy

\bibliographystyle{apalike}\biboptions{authoryear}

\begin{frontmatter}

\title{Generative Design of Physical Objects using Modular Framework}

%Group authors per affiliation:
\author{Nikita O. Starodubcev}
\ead{nstarodubtcev@itmo.ru}
\author{Nikolay O. Nikitin}
\ead{nnikitin@itmo.ru}
\author{Konstantin G. Gavaza}
\author{Elizaveta A. Andronova}
\author{Denis O. Sidorenko}
\author{\\ Anna V. Kalyuzhnaya}

\address{ITMO University, Saint-Petersburg, Russia}

\begin{abstract}
In recent years generative design techniques have become firmly established in numerous applied fields, especially in engineering. These methods are demonstrating intensive growth owing to promising outlook. However, existing approaches are limited by the specificity of problem under consideration. In addition, they do not provide desired flexibility. In this paper we formulate general approach to an arbitrary generative design problem and propose novel framework called GEFEST (Generative Evolution For Encoded STructure) on its basis. The developed approach is based on three general principles: sampling, estimation and optimization. This ensures the freedom of method adjustment for solution of particular generative design problem and therefore enables to construct the most suitable one. A series of experimental studies was conducted to confirm the effectiveness of the GEFEST framework. It involved synthetic and real-world cases (coastal engineering, microfluidics, thermodynamics and oil field planning). Flexible structure of the GEFEST makes it possible to obtain the results that surpassing baseline solutions.

\end{abstract}


\end{frontmatter}

\section{Introduction and problem definition}

Over the past decades artificial intelligence, machine learning, and optimization methods have become an inevitable part of the solution of engineering design problems \citep{chen2020heat, steinbuch2010successful, zheng2021generative}. These methods demonstrate great potential to improve and simplify tasks usually performed by engineers. Particular complexity of engineering design problems is associated with large design space originated by a great number of parameters \citep{danhaive2021design, harding2016dimensionality}. Human efforts are not enough to explore such high dimension space. In contrast, computational based approaches can be examined as an efficient tool for given purposes. 

The most common computational based methods to an engineering design problem are generative design \citep{vajna2005autogenetic} and topology optimization \citep{bendsoe1989optimal}. In general, the main goal of these methods is the same - to find one or several physical objects whose properties more preferable than those of existing ones taking into account its geometrical and boundary restrictions \citep{bendsoe1988generating, bendsoe1989optimal, vajna2005autogenetic}. The key difference lies in the way this goal is reached.  Topology optimization seeks to enhance performance and reduce the weight of the already existing objects via optimizing material distribution in it \citep{bendsoe1989optimal, tyflopoulos2018state}. Conversely, in generative design there is no prior knowledge about initial object, it "generates" structures based on space constraints and design goals only \citep{vlah2020evaluation}. Aforementioned approaches have gained widespread acceptance in various applied fields, for instance, ocean engineering \citep{tian2022optimization}, mechanical design \citep{oh2019deep}, heat and mass transfer \citep{qian2022adaptive}, thermal engineering \citep{zou2022topology}. However, only for topology optimization well-defined theory fundamentals were established \citep{bendsoe1988generating, bendsoe1989optimal}, whereas strict problem statement in generative design of physical objects is still missing \citep{vlah2020evaluation}.

In previous works several definitions of generative design were proposed, \cite{shea2005towards}: “Generative design systems are aimed at creating new design processes that \textbf{produce} spatially novel yet efficient and buildable designs through exploitation of current computing and manufacturing capabilities”, \cite{kallioras2020dzai}: 
"Generative Design is the methodology for automatic \textbf{creation} of a large number of designs via an iterative algorithmic framework while respecting user-defined criteria and limitations". Nevertheless, the production and creation mechanisms of designs were not explained accurately. Originally, in statistical theory, generative modelling problem have been aimed at reconstruction the joint probability distribution $\mathbf{P}(\mathbf{X, Y})$ on random variables $\mathbf{X}$ (observable variable) and $\mathbf{Y}$ (target variable) \citep{ng2001discriminative, jebara2012machine}. Generative design is associated with the same problem, but joint distribution is extremely intricacy. Special complexity is related to the variables $\mathbf{X}$ and $\mathbf{Y}$ that correspond to a real physical object and its performance, respectively. For instance, heat-generating components of electronic systems and temperature field \citep{qian2022adaptive}, car wheel and its cost, novelty, compliance \citep{oh2019deep}, ship hull form and its strength \citep{liu2022multi}. In generative design, objects with the highest performance are of greatest interest, and joint distribution allows to obtain (produce or create) such physical objects using sampling procedures. Thus, the \textbf{production} and \textbf{creation} mechanisms of designs lies in sampling from $\mathbf{P}(\mathbf{X, Y})$, and generative design problem is focused on the estimation of joint probability distribution on real physical objects and its performance.

However, reconstruction of $\mathbf{P}(\mathbf{X, Y})$ for real objects out of reach to date owing to: 1) extremely high dimension of design space; 2) numerous geometrical and boundary restrictions on $\mathbf{X}$ variable; 3) target space is frequently continuous and might be multi-dimension. Existing approaches enable to obtain only minority of samples from joint distribution. 
\begin{figure*}[t!]
\centering
\includegraphics[width=16cm]{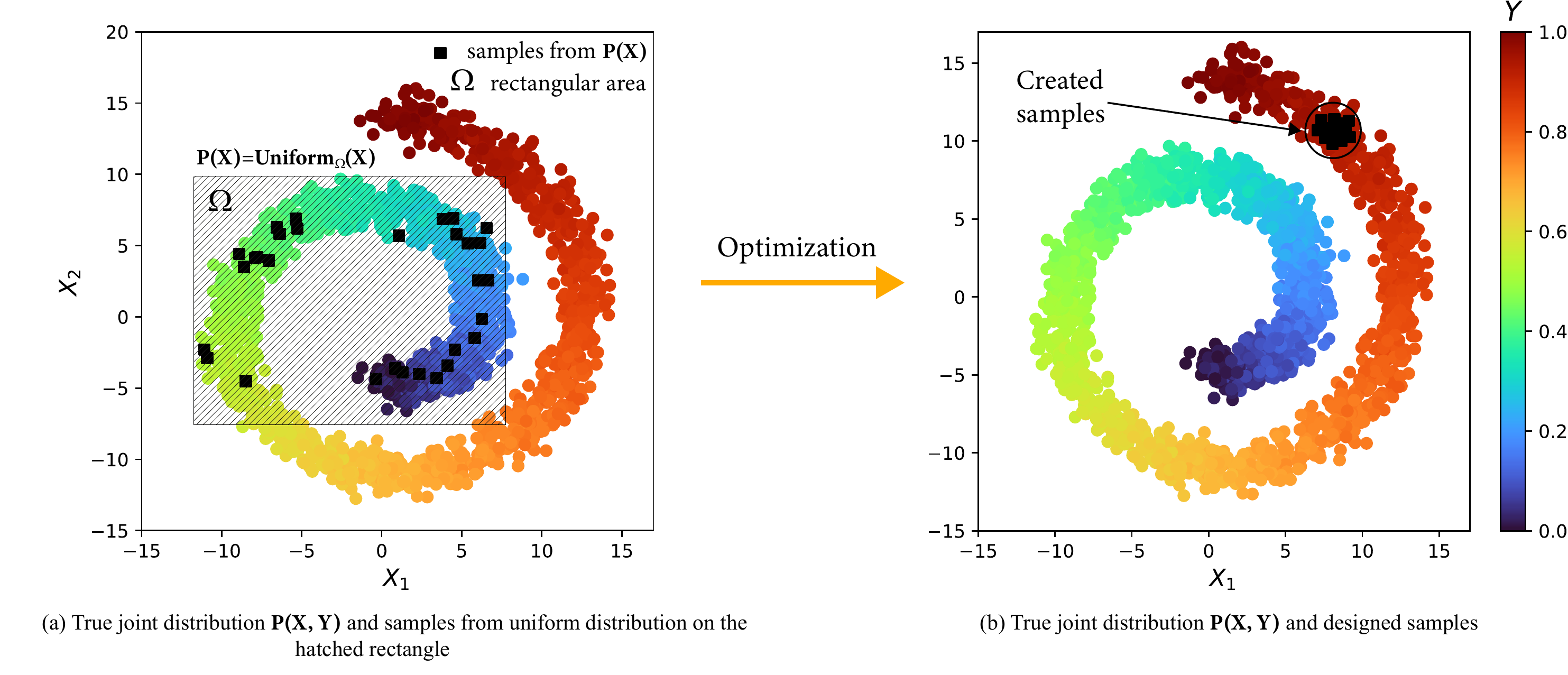}
\caption{Two-dimensional example $\mathbf{X}=(\mathbf{X_{1}, X_{2}})$ for sample production from $\mathbf{P}(\mathbf{X, Y})$ using uniform distribution, gradient boosting and genetic algorithm as object distribution, conditional target distribution and optimization method, respectively.}
\label{example}
\end{figure*}
All of them include following stages (we call this a \textit{generative design procedure}\label{procedure}): 
\begin{enumerate}
    \item to define object distribution  $\mathbf{P}(\mathbf{X})$ to sample $\mathbf{X}$;
    \item to model conditional target distribution $\mathbf{P}(\mathbf{Y}|\mathbf{X})$ to acquire performance of sampled $\mathbf{X}$;
    \item to solve optimization problem
\begin{equation} \label{optim}
    \begin{gathered}
        \mathbf{X}^{*} = \argmaxA_{\mathbf{X}\sim\mathbf{P}(\mathbf{X})}\mathbf{Y}(\mathbf{X})
    \end{gathered}
\end{equation}
\end{enumerate}
Implementation example of mentioned procedure used for sample creation from joint distribution $\mathbf{P}(\mathbf{X, Y})$ in two-dimensional space is shown in Figure \ref{example}. Uniform distribution, gradient boosting and genetic algorithm were used as object distribution $\mathbf{P(X)}$, conditional target distribution $\mathbf{P}(\mathbf{Y|X})$ and optimization method, respectively. It can be clearly seen that the number of designed $\mathbf{X}$ is low compared to the number of all possible samples from $\mathbf{P}(\mathbf{X, Y})$. Detailed discussion of outlined stages is given in Section \ref{sec_related}. 

In this paper we propose flexible open-source framework GEFEST (Generative Evolution For Encoded STructures) for generative design of two-dimensional physical objects from various applied fields. We consider physical objects that can be represented as polygons neglecting their internal structure. Our approach is based on the \textit{generative design procedure}.  Flexibility of the framework is attained due to possibility of toolkit construction for particular applied problem, where toolkit implies the set of methods for implementation of the \textit{generative design procedure}. Since the GEFEST framework offers different approaches for each stage, it is possible to select the most suitable of them for considered problem and thereby to improve the final designs. Moreover, we provide an opportunity to implement custom approaches and modify the GEFEST core. In addition, the framework allows to operate with physical objects of different nature due to universal representation of each processed object. Our main contributions are the following: 
\begin{itemize}
    \item Formulation of the general approach to an arbitrary generative design problem.
    \item Novel generative design framework implemented as an open-source tool. The novelty lies in the individual approach to the solution of each stage in the \textit{generative design procedure} achieved through flexible combination and modification of multiple methods for particular problem.
    \item Validation of our framework on several real-world problems and comparison with baselines.
\end{itemize}

The paper is organized as follows. In Section \ref{sec_related} we consider related works to this paper. In Section \ref{sec_approach} we explain our general approach to generative design problem. In Section \ref{sec_software} we describe implementation details of proposed framework. In Section \ref{sec_exp} we show real-world applications of the GEFEST framework. Finally, in Section \ref{sec_disc} and Section \ref{conclusions} we conclude this work with pros ans cons of our framework and future directions of research.

\section{Related works}
\label{sec_related}
In this section a review of different approaches to the solution of each stage in the \textit{generative design procedure} step will be taken. Moreover, the existing generative design frameworks will be discussed.
\subsection{Definition of object distribution}
Brute force approach involves the selection of standard distributions $\mathbf{U(X)}$, for example, uniform or normal as $\mathbf{P(X)}$ \citep{nikitin2021generative, mukkavaara2020architectural}. $\mathbf{U(X)}$ is defined on a variable $\mathbf{X}$ from design space $\mathcal{D} = \{ \, \mathbf{ X} \in \mathcal{\mathbb{R}}^{n}\, |\, f(\mathbf{ X}) = 1 \, \}$, where $f = \{ 0, 1\}$ is geometrical and boundary constraint identification function, $n \, - $ dimension of design variable $\mathbf{X}$. Described method is accompanied by several challenges. Firstly, dimension $n$ may vary within one considered problem, what causes difficulties in further conditional distribution $\mathbf{P(Y|X)}$ modelling. Secondly, sampling procedure from $\mathbf{U(X)}$ will be time-consuming if form of $\mathcal{D}$ is non trivial. In other words, samples $\mathbf{X}\sim\mathbf{U(X)}$ may not satisfy the boundary conditions, therefore it should be rejected by $f$, and sampling algorithm should be repeated afterwards. Lastly, diversity of samples can be poor because of the simplicity of the selected standard distribution.

Another rapidly developing class of approaches for $\mathbf{P(X)}$ estimation is based on deep generative neural networks \citep{oh2019deep, qian2022adaptive, tan2020deep}. These are data-driven approaches, usually using implicit (or semi-implicit) models. The latter work in black box mode, and are able only for generating samples $\mathbf{X}$. In addition, they commonly require a great amount of data as well as training time. Despite the shortcomings, if the models are well trained, this class will be free from challenges that were inherent in the brute force approach. The vast majority of approaches are based on generative adversarial networks \citep{goodfellow2014generative} and variational autoencoders \citep{kingma2013auto}.

\subsection{Modelling of conditional target distribution}
In generative design a well-established approach to the performance estimation (or estimation of the conditional target distribution $\mathbf{P(Y|X)}$) of objects is numerical modelling. It is models based on equations of mathematical physics that can be solved using physical simulators $\mathbf{Sim(X)}$, for instance, species distribution modeling \citep{xu2021ecological}, computational fluid dynamic \citep{xu2021ecological}, COMSOL multiphysics \citep{nikitin2021generative}, Simulating WAves Nearshore \citep{james2018machine}. Despite %such
equation-based models provide highly accurate approximation of performance (i.e. $\mathbf{Sim(X)} \approx \mathbf{Y}$), this approach is computationally expensive and consequently extremely time-consuming.

Another approach that focuses on solving the mentioned problem of high computational complexity is surrogate models \citep{chen2020heat, palar2019use, gonzalez2016optimisation}. The main idea is to build a lightweight data-driven model $\mathbf{Surr(X)}$ that will approximate outputs of  $\mathbf{Sim(X)}$ with reasonable accuracy. A wide range of machine learning and deep learning methods can be used as a surrogate model: random forest, kriging, gradient boosting, neural networks, etc.

\subsection{Solving optimization problem}
After $\mathbf{P(X)}$ and $\mathbf{P(Y|X)}$ are specified, the last stage of the \textit{generative design procedure}, i.e., optimization problem (\ref{optim}), should be discussed. This is the main part to get samples with the highest performance from joint distribution $\mathbf{P(X, Y)}$, which is the goal of generative design. Gradient based methods and biologically inspired algorithms can be considered as tools for this purpose.

In generative design different variations of evolutionary algorithms are the most common approaches in solving the optimization problem for real physical objects \citep{shen2022metamodel, qian2022adaptive, nikitin2021generative}. The increased attention to emphasized algorithms caused by the following reasons: 1) gradients of $\mathbf{Sim(X)}$ with respect to $\mathbf{X}$ are hardly calculated, whereas evolutionary algorithms are gradient free; 2) evolutionary algorithms can be easily generalized to multi-criteria optimization problem ($\mathbf{Y} \in \mathbb{R}^{k}, k>1$), while for gradient based methods this operation is more complicated. However, the convergence of such algorithms strongly depends on genetic operators and in some cases can be time-consuming. 

With the growth of machine learning technologies, gradient approaches are gaining more spread in generative design \citep{tan2020deep}. If the surrogate model ensures a high approximation accuracy, in the optimization problem it is possible to replace $\mathbf{Sim(X)}$ with a $\mathbf{Surr(X)}$. Modern optimizers, such as Adam or RMSProp, enable to obtain gradients of machine learning model $\mathbf{Surr(X)}$ quickly and efficiently.

\subsection{Generative design frameworks}
Attempts to develop generative design frameworks have been made for quite some time. It is necessary to highlight the work presented by \cite{singh2012towards}. In this paper authors proposed framework based on different generative design approaches: genetic algorithms, swarm intelligence, L-systems, cellular automata and shape grammars. After all, they came to the conclusion that there is no universal approach to the generative design problem. In other words, a generative design framework needs to be flexible, that is, provides various approaches to a specific problem. 

In recent years the development of generative design frameworks had become more widespread. For example, \cite{mukkavaara2020architectural} devised a framework for architectural design. In the presented paper researchers tried to develop generic framework including several generators of designs (genetic algorithm and random sampling). However, this work suffer from the lack of modern deep learning models. 

Moreover, the Autodesk generative design framework \citep{buonamici2020generative} deserves special attention as one of the best-known commercial software. This framework provides flexible approach for user-specified problem. For obtaining designs numerical analysis of differential equations is performed on external cloud servers. Despite the advantages of the given software, its implementation details and essential features are not discussed.

Increasingly, works with a combination of deep learning networks and traditional approaches are being published. For instance, \cite{oh2019deep} presented a framework which consists of topology optimization (Solid Isotropic Material with Penalization) and generative model (Generative Adversarial Network). This approach demonstrated high diversity and aesthetics of created designs in resolving two-dimensional wheel problem. Nevertheless, it has not been validated on tasks from different applied fields.

\section{GEFEST approach for generative design}
\label{sec_approach}

As was mentioned earlier there is no universal approach to the generative design. Each applied problem requires detailed consideration and long-term research. However, the combined application of deep learning, numerical modelling and optimization can be regarded as a general trend in solution of the generative design problem. With the right combination of different methods from these specified branches, well-performed physical objects can be produced. This is the main idea of GEFEST approach.

\begin{figure*}[h!] 
    \centering
    \includegraphics[width=16.3cm,height=7.8cm]{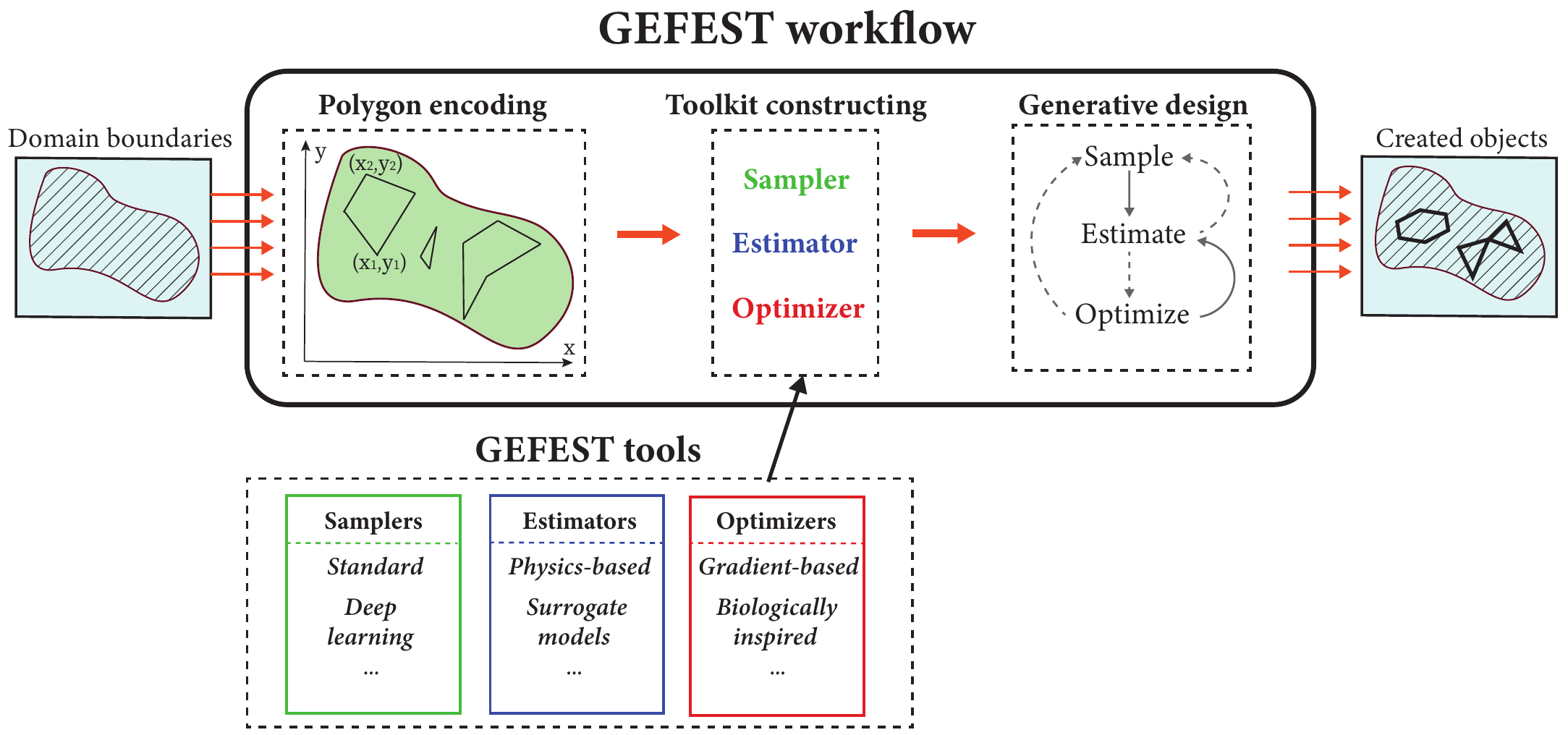}
    \caption{The GEFEST approach for the generative design of physical objects including polygon encoding, toolkit constructing and generative design. In generative design stage hatched and thick line means possible and required gates, respectively.}
    \label{workflow}
\end{figure*}

The GEFEST framework is a modular tool for the generative design of two-dimensional physical objects. The essential points of the framework architecture are shown in Figure~\ref{workflow}. First of all, the GEFEST receives the boundaries of considered two-dimensional domain as input. Optimization will be carried out in this domain. Then workflow including polygon encoding, toolkit constructing and generative design is performed. At the output, the GEFEST generates samples from the joint probability distribution $\mathbf{P(X, Y)}$. 
\subsection{Polygon encoding}
\begin{figure}[H]
    \centering
    \includegraphics[width=9cm]{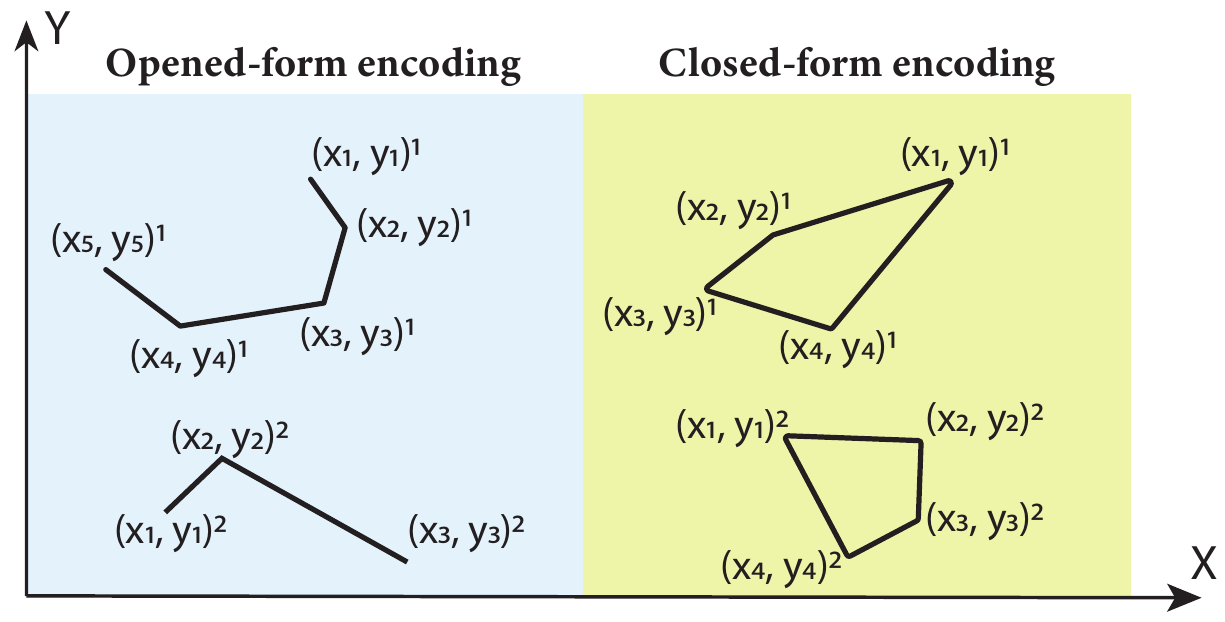}
    \caption{Opened-form and closed-form polygon encoding in Cartesian coordinates}
    \label{example_poly}
\end{figure}

The first stage of the GEFEST workflow is polygon encoding. This procedure allows to represent real physical objects as two-dimensional (flat) polygons. We distinguish two types of structure: opened-form and closed-form, nevertheless framework operates them in the same manner. An example of encoding is shown in Figure \ref{example_poly}. 
It is clear that each polygon node is identified by a corresponding Cartesian coordinates. Hence, the polygon can be described by a set of points, i.e. $\mathbf{X} = (x_1, y_1, ..., x_j, y_j), \mathbf{X} \in \mathbb{R}^{2j}$. If $\mathbf{X}$ satisfies the boundary and geometrical constraints ($f = 1$), it will belongs to the design space $\mathcal{D}$. The restrictions are based on the domain boundaries that are obtained as input in the GEFEST. Thus, design space is defined by all possible polygons $\mathbf{X}$ and their combinations. It is worth noting that the dimension of this space is high, especially for closed-form polygons.

\subsection{Toolkit constructing using GEFEST tools}
After the specification of the design space, it is necessary to set a certain approach for each stage of the \textit{generative design procedure}. To accomplish this purpose, we have implemented GEFEST tools (\textit{Samplers}, \textit{Estimators}, \textit{Optimizers}) including various computational methods. Semantically, all tools can be divided into two classes: deep learning (Generative Adversarial Networks, Variational Auto Encoders, etc) and standard (standard statistical distributions) for Samplers; surrogate models (fully connected/convolutional neural networks, kriging, etc) and physics-based (different physics simulators) for Estimators; biologically-inspired (genetic/evolutionary algorithms, etc) and gradient-based (Adam, gradient descent, etc) for Optimizers. Such variety of approaches provides the opportunity to deal with applied problems of different nature and goals. For example, when solving a problem with opened-form polygons and multi-criteria target \citep{nikitin2020multi}, it will be effective to select a standard distribution and  multi-objective evolutionary algorithm as sampler and optimizer, respectively. In cases requiring high diversity of polygon with closed-form, deep learning sampler is more preferable.

\subsubsection{GEFEST standard sampler}
\begin{figure*}[h!]
    \centering
    \includegraphics[width=16cm]{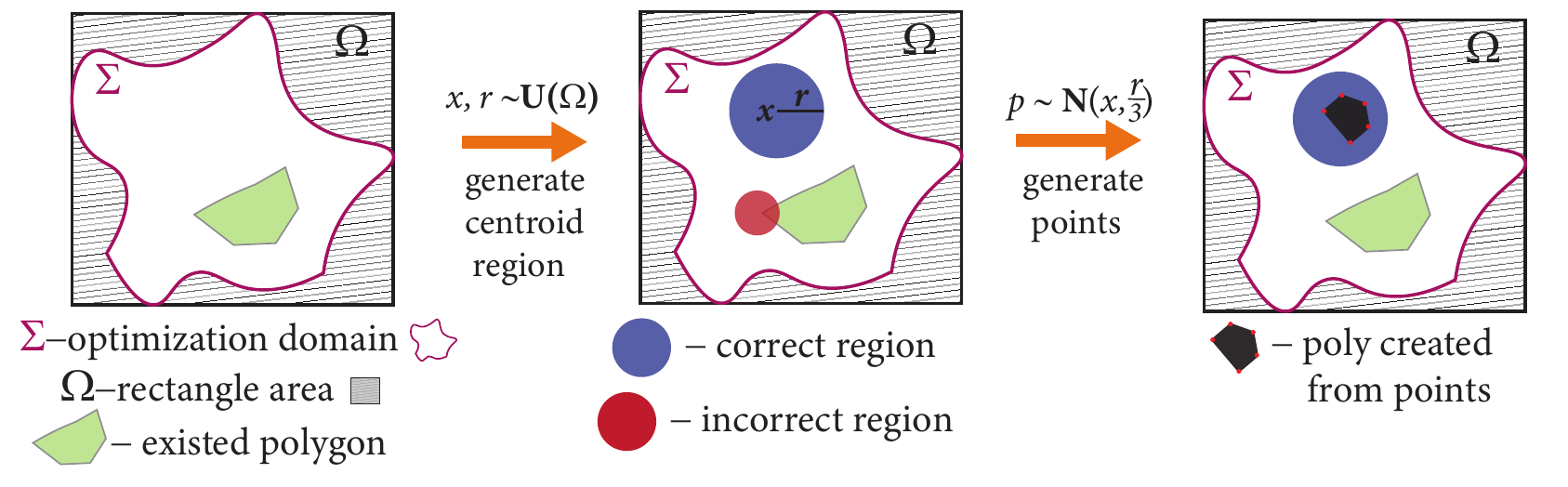}
    \caption{Visualization of single step of the GEFEST standard sampler. Our sampler operates in two main stages: centroid generation and points generation inside the centroid region. Here $\mathbf{U}(\Omega)-$ uniform distribution on rectangle area; $x, r \ -$ centroid and radius of the centroid region, respectively; $\mathbf{N}(x, \frac{r}{3})-$ normal distribution, where $x, \frac{r}{3}$ - its mean and variance, respectively.}
    \label{standard_sampler}
\end{figure*}

The most principal requirement imposed on the sampler is computational efficiency of sample generation. We would like to create correct polygons, i.e. without any self-intersections, out-of-bound parts and intersection with other domain elements, in an acceptable time. For these purposes, we have implemented GEFEST standard sampler including two stages: generation of the centroid region and generation of points inside this region. We refer our sampler to the standard class because it is based on standard statistical distributions. In a simplified form sampling procedure is presented in Algorithm \ref{standard_algo} and Figure \ref{standard_sampler}.
\begin{algorithm}[h!]
\caption{The GEFEST standard sampling}\label{standard_algo}
\begin{algorithmic}[1]
\Require $\Omega, \Sigma, N, max$ \Comment{Rectangle area, optimization domain, number of polygons to generate and maximum number of points in polygon}
\Ensure $S$ \Comment{The GEFEST Structure (set of polygons)}
\State $S = Array()$
\While{$|S| < N$ }
\State $n_{poly} = |S| + 1$ \Comment{Number of already created polys and additional one to avoid 0}
\State $x \sim \mathbb{U}(\Omega)$ \Comment{Creating centroid}
\While {$x$ not in $\Sigma$}
\State $x \sim \mathbb{U}(\Omega)$ \Comment{Repeat until $x$ is in optimization domain}
\EndWhile
\State $r \sim \mathbb{U}((0, \frac{|\Omega|}{n_{poly}}])$ \Comment{Creating radius of centroid region}
\While {$\vec{xr}$ \textit{is incorrect}}
\State $r \sim \mathbb{U}((0, \frac{|\Omega|}{n_{poly}}])$ \Comment{Repeat if region is incorrect}
\EndWhile
\State $n_{point} = \mathbb{U}_{int}(max)$ \Comment{Number of points}
\State $P = Array()$ \Comment{Initialization of polygon}
\While{$|P| \neq n_{point}$}
\State $p \sim \mathbb{N}(x, \frac{r}{3})$ \Comment{Creating polygon point}
\State $P.append(p)$
\EndWhile
\State $S.append(P)$
\EndWhile
\State \Return $S$
\end{algorithmic}
\end{algorithm}

First of all, the centroid is created using uniform distribution on rectangle area. This poses the central point of the region called centroid region. The size of the latter is determined by the radius, which is also sample from uniform distribution. However, the support of this new one is different. More precisely, for the radius we considered uniform distribution on the ray $(0, \frac{|\Omega|}{n_{poly}}]$ instead of rectangle area $\Omega$. Selection of such upper bound is motivated by the following idea: when the number of polygons increases, it becomes more difficult to find a correct region with a large radius. To avoid this, the ray can be reduced by $n_{poly}$ times. Finally, when the correct region is created, a polygon can be freely generated inside it. The polygon consists of points sampled from normal distribution with parameters $x$ and $\frac{r}{3}$, the mean and the variance, respectively (the latter is chosen taking into account the three-sigma rule). Note that in Algorithm \ref{standard_algo} we do not adduce details about stopping criterion (when the while loop takes a good deal of time), postprocessing and recreating the centroid (in cases when radius creation becomes time-consuming). 

\subsubsection{Deep learning estimators and samplers}
Within the GEFEST framework, deep learning estimator and sampler works with images of polygons, not the polygons proper. This is caused by the various dimension of the latter. As can be seen from Figure~\ref{example_poly}, the number of points describing the polygons can be different. It depends on the number of its segments. Thus, in the application of classical machine learning methods certain difficulties arise (dimension of the input data varies). To avoid these issues, it is necessary to make an universal polygon parametrization insensitive to the number of its points. In this work we have considered three-dimensional matrix parametrization invariant to changes in the number of points. In other words, an image is mapped to each polygon or polygon structure. The produced images are binary in which maximum intensity corresponds to the polygon. Such representation allows to consider a well-established practical tools, namely convolutional neural networks.

\begin{figure}[h!]
    \centering
    \includegraphics[width=9cm]{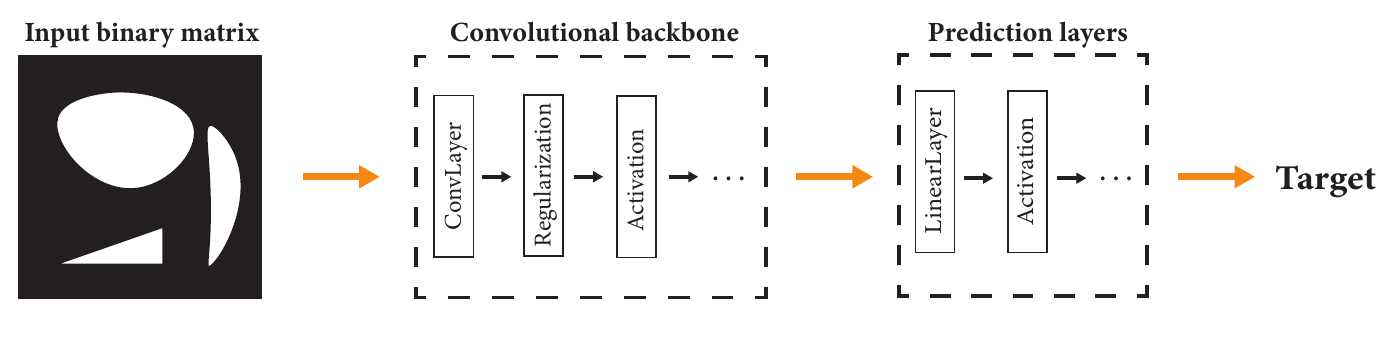}
    \caption{The generalized architecture of the deep learning estimator takes the image of polygons as input.}
    \label{estimator_arch}
\end{figure}

The generalized architecture of the deep learning estimator is shown in Figure~\ref{estimator_arch}.
The estimator passes an input binary matrix through convolutional backbone and prediction layers to approximate target. As options of backbone, %traditional
various widely-used convolutional architectures \citep{khan2020survey} can be listed: VGG, ResNet, UNet, AlexNet. The choice is dictated by the complexity of considered problem and the amount of training data. Fully connected networks are usually used as prediction layers. 

\begin{figure}[h!]
    \centering
    \includegraphics[width=9cm]{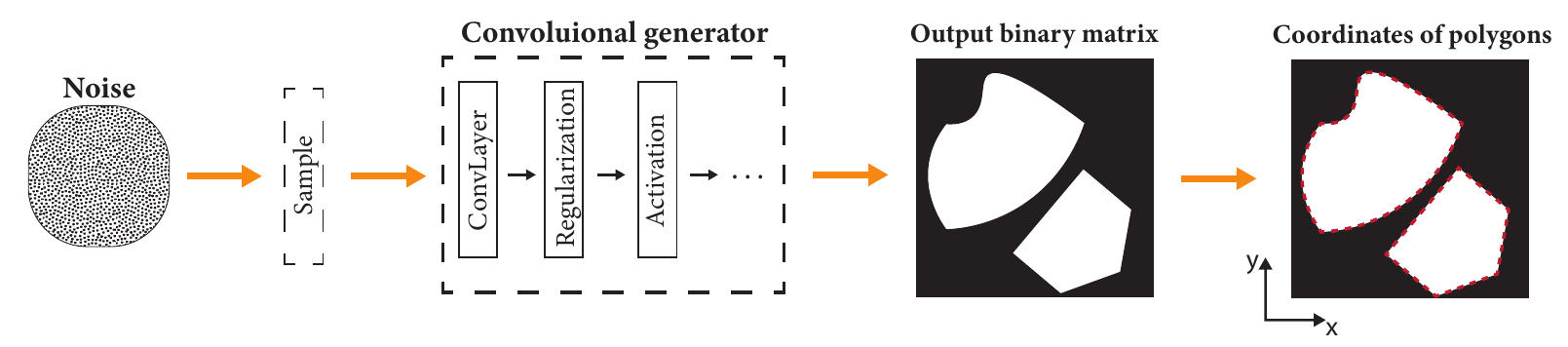}
    \caption{The generalized architecture of the deep learning sampler in the GEFEST. It tries to produce a sample from the noise distribution using convolutional generator.}
    \label{sampler_arch}
\end{figure}

Deep learning sampler works the opposite way, it tries to create a realistic output binary matrix as shown in Figure~\ref{sampler_arch}.
The main purpose of the sampler is to create plausible binary matrix from noise. The generator should produce images with the correct geometric form of polygons. The most common approaches to the construction of a convolutional generator are Variational Auto Encoder, Generative Adversarial Networks, Normalizing Flows, etc. The final step is the transformation from matrix parametrization to Cartesian coordinates. This can be done using classical computer vision algorithms to detect edges \citep{canny1986computational}.

\subsubsection{Evolutionary core}\label{core}
\begin{figure}[h!]
    \centering
    \includegraphics[width=9cm]{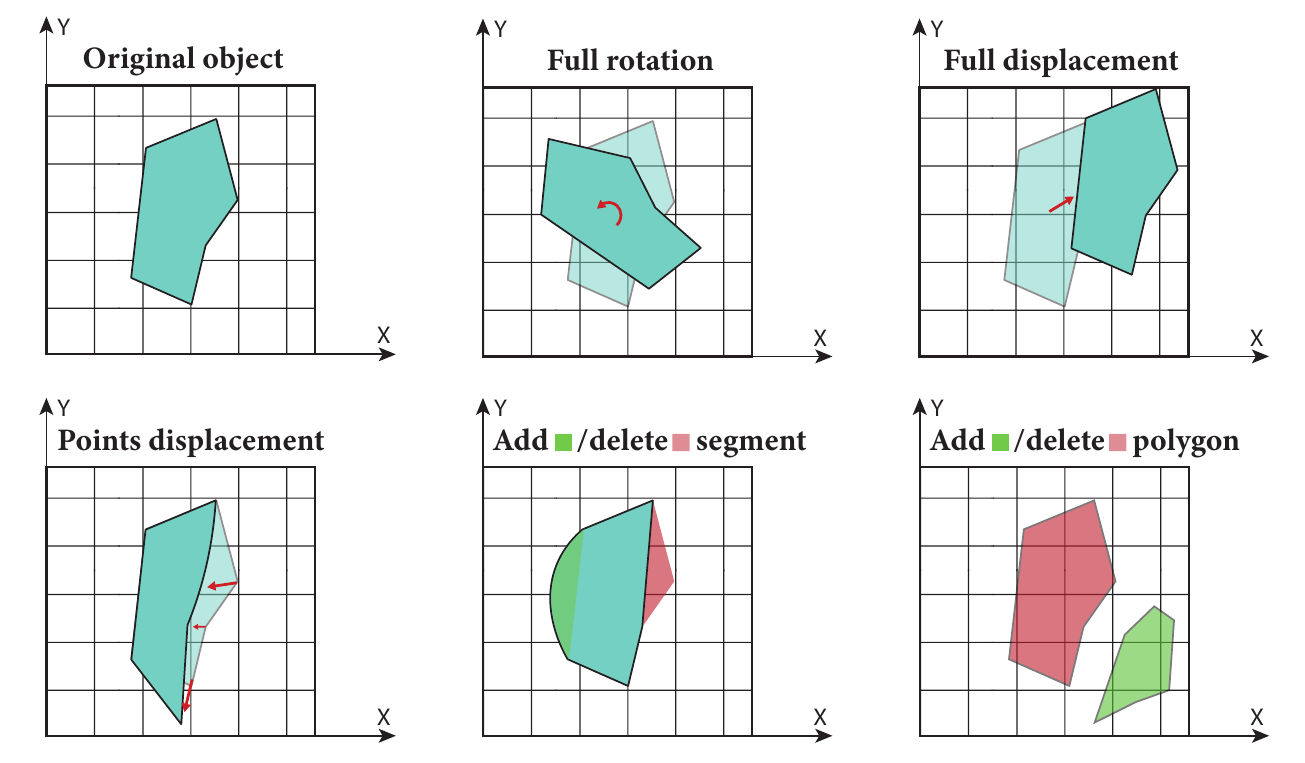}
    \caption{Geometrical transformation over polygons: rotation, displacement, adding, removal}
    \label{mutat}
\end{figure}

\begin{figure}[h!]
    \centering
    \includegraphics[width=9cm]{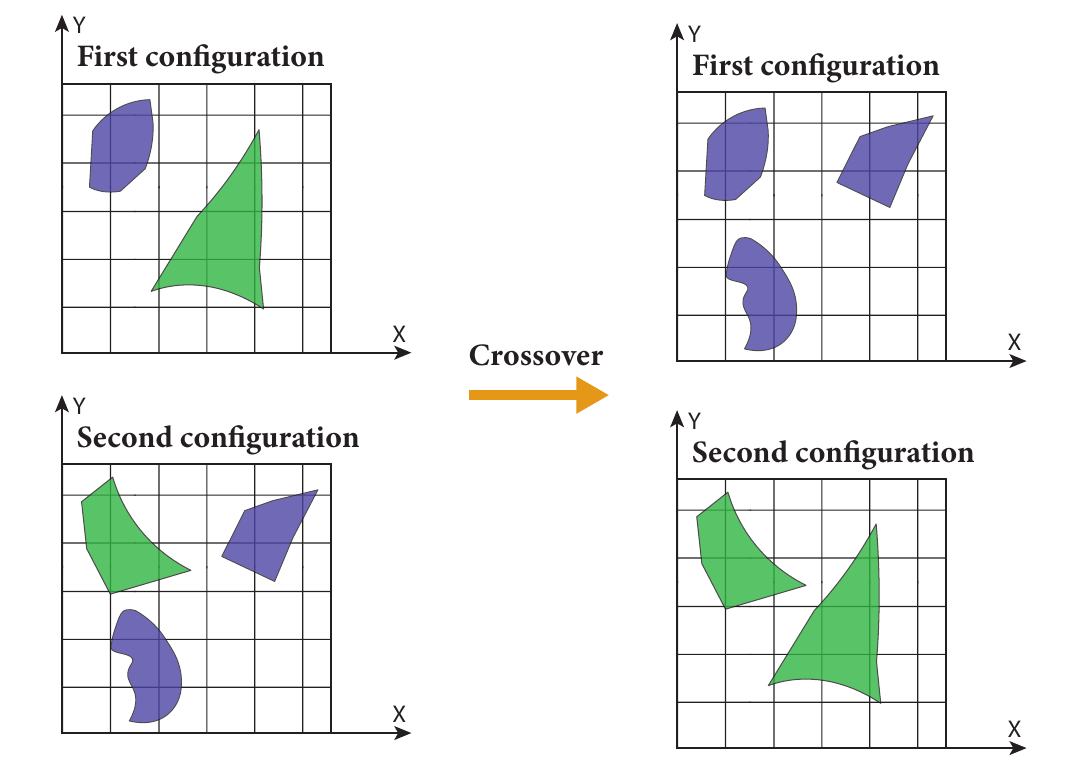}
    \caption{Open-form and closed-form polygon encoding in Cartesian coordinates}
    \label{cross}
\end{figure}

The optimization step, particularly biologically-inspired, deserve special attention. It is known that convergence rate of every evolutionary algorithm is strongly depends on the genetic operators \citep{SongJ021}. For effective convergence, the latter must be implemented taking into account the semantics of the problem being solved. In case of two-dimensional polygons, it is natural to consider geometric transformations shown in Figure \ref{mutat} as mutation operators. 
We have realized following transformations over polygons: rotation of the center of mass by certain angle; polygon and points displacement; adding/deleting segments and polygons. Moreover, we have used crossover operator shown in Figure \ref{cross}.
These genetic operators constitute an evolutionary core of the GEFEST. More precisely, every evolutionary algorithm used in the framework utilizes this evolutionary core.

\subsection{Generative design}
The fundamental stage of the GEFEST workflow is the generative design, depicted as last step in Figure \ref{workflow} and in Algorithm \ref{gen_des}.
\begin{algorithm}[h!]
\caption{GEFEST generative design}\label{gen_des}
\begin{algorithmic}[1]
\Require $P = \left(S, E, O \right)$ \Comment{User-defined toolkit}
\Ensure $D_{fin}$ \Comment{Final designed objects}
\State $D_{curr} \gets S.\text{sample()}$ \Comment{Initial designs}
\While{$stopCriteria$}
\State $PF \gets E.\text{estimate}(D_{curr})$ \Comment{Design performance}
\State $D_{curr} \gets E.\text{select}(PF, D_{curr})$ \Comment{Selecting $k$ best objects}
\If{$O.\text{required}$}
    \State $D_{curr} \gets O.\text{optimize}(D_{curr}, PF)$
    \If{$S.\text{required}$}
        \State $D_{sample} \gets S.\text{sample()}$
        \State $D_{curr} \gets D_{sample} \cup D_{curr}$ \Comment{Combination}
    \EndIf
\Else{} \Comment{Skip optimization}
    \State $D_{sample} \gets S.\text{sample()}$
    \State $D_{curr} \gets D_{sample} \cup D_{curr}$
\EndIf
\EndWhile
\State $D_{fin} \gets D_{curr}$
\State \Return $D_{fin}$
\end{algorithmic}
\end{algorithm}
This procedure is based on three principles: sampling, estimation, optimization, which can be combined in different ways. The \textit{traditional approach} includes lines 1, 3, 4 and 6 in Algorithm \ref{gen_des}. More exactly, it performs single sample operation, whereas estimation and optimization are repeated until the stopping criterion is reached. Therefore, low exploration and high exploitation rate are inherent in this process. However, in some problems it is necessary to increase the exploration rate. This can be done by integration of lines 8 and 9, i.e. by executing of addition sample operation at each stage of the loop, we call this \textit{extra sampling}. On the other side, it is possible to skip optimization phase completely (lines 1, 3, 4, 11 and 12). Such a method, characterized by great exploration rate, is called \textit{random search}.

\section{Open-source software framework}
\label{sec_software}
\begin{figure*}[h!]
    \centering
    \includegraphics[width=16cm]{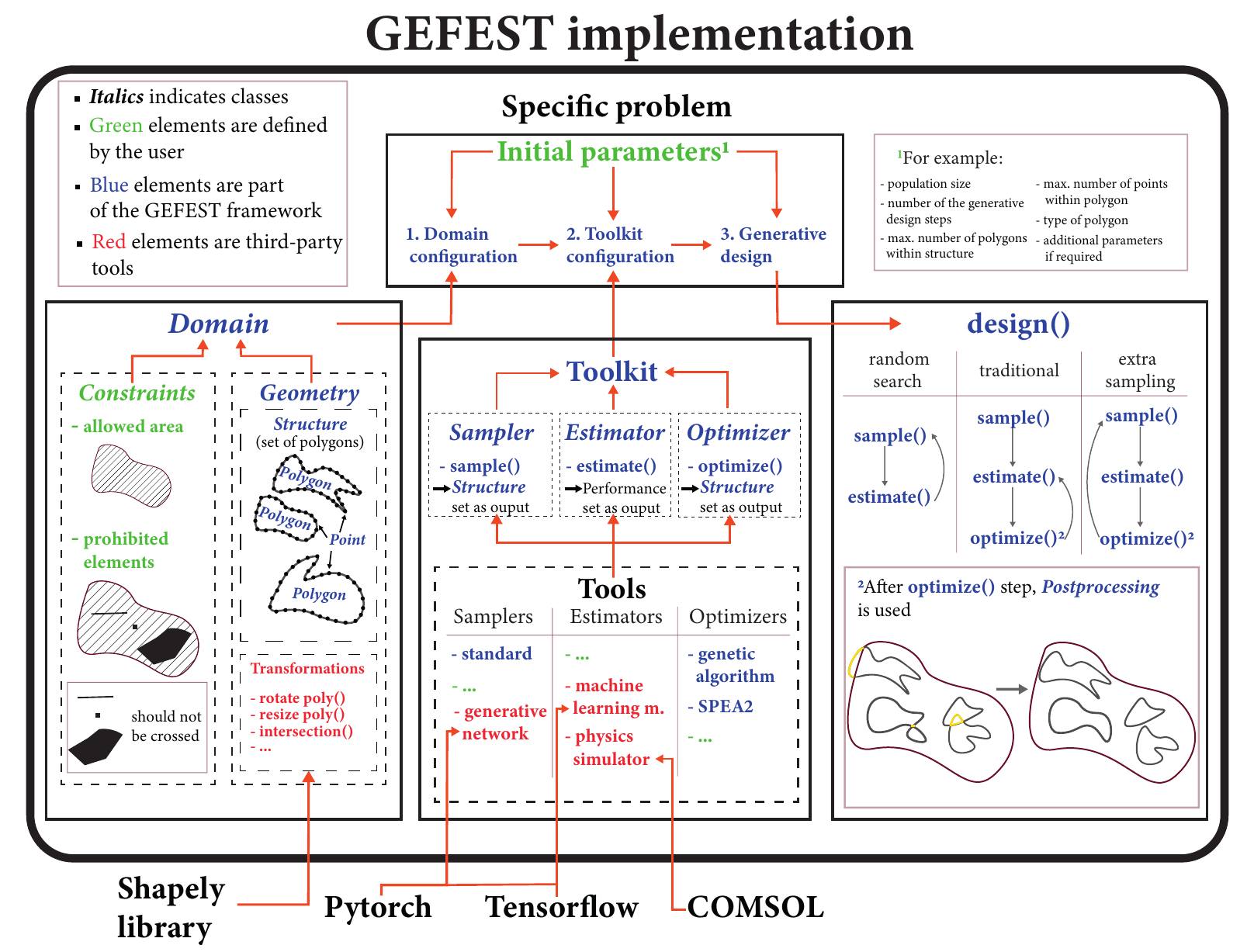}
    \caption{The scheme of the GEFEST framework implementation. The specific problem can be solved by completing three main steps: domain and toolkit configuration and design.}
    \label{fig_class}
\end{figure*}

We provide our framework as an open-source tool for solution of the user-specified generative design problem. The architecture of the GEFEST framework is presented in Figure~\ref{fig_class}. A certain problem may be solved by adjustment of the following main blocks: domain, toolkit and design. Worth noting that access to them must be carried out in the presented order. This follows from the fact that subsequent blocks require elements configured at the previous stages. Furthermore, to make available user interaction, we divided GEFEST elements into three groups: user-defined, internal and external.

\subsection{Domain}
First of all, it is necessary to configure \textit{Domain} class, which is responsible for the whole information about geometry of the problem. In order to do this, the user should  define \textit{Constraints}, namely the allowed area and prohibited elements. The first of them determines the domain in which optimization will take place. The second one answers for fixed elements within the domain that should not be intersected by generated polygons. 

In addition to the \textit{Constraints}, we set the \textit{Geometry} class, which consists of \textit{Structure} and transformations. The \textit{Structure} is necessary for creation of abstraction over real objects in accordance with the following hierarchy scheme: \textit{Structure} $\rightarrow$ \textit{Polygon} $\rightarrow$ \textit{Point}. Lastly, to realize the geometrical transformations over polygons (for instance, rotation, resizing and etc) we provide an access to the external methods from the Shapely library \citep{shapely}.
 
\subsection{Toolkit}
The next stage, after the \textit{Domain} specification, lies in toolkit configuration. Notes that this is the core part affecting the performance of created objects. The toolkit comprises three classes: \textit{Sampler}, \textit{Estimator} and \textit{Optimizer} realizing corresponding abstract methods (sample(), estimate() and optimize()). Such an abstraction is necessary to define the general behavior of objects, which are insert in the GEFEST tools. In the latter block, we implemented our own objects and external elements from machine learning frameworks and physics-based applications. In addition, we provided access to custom tools.

It should be pointed out that the main requirement imposed on tools is consistency. In other words, inputs and outputs of class methods should have the same type. For example, if the sample() method delivers a \textit{Structure} array, then the input of the estimate() method should have the same type. Initially, all methods operate with the \textit{Structure} array, however other options are available (for example, array of images).

\subsection{Design}
The final step is a generative design based on Algorithm \ref{gen_des}. Here we effort an opportunity to three options (random search, traditional and extra sampling methods), as was discussed earlier. We just emphasize that after the optimize() step, \textit{Postprocessing} is utilized. This means that defective polygons (self-intersected, out-of-domain and etc) will be corrected via the postprocessing.

\section{Experimental studies}
\label{sec_exp}
In this section the application of GEFEST framework to physical objects of different nature is considered. The main purpose of the experiments lay in the demonstration of framework versatility by addressing the real-world problems from different specific areas using the same GEFEST approach (Figure \ref{workflow}). As it was mentioned earlier, such beneficial feature stems from flexible toolkit construction caused by various options for each stage of the \textit{generative design procedure}. In order to provide practical suggestions on the GEFEST core modification to improve the effectiveness of solutions, we have realized several combinations of different methods. Moreover, we have validated our framework to reveal its limitations. Summary of all experimental studies is presented in Table \ref{tab_experiments}
\begin{table}[h!]
\caption{\label{tab_experiments} Summary of all experimental studies including the main goal.}
\resizebox{\columnwidth}{!}{%
  \begin{tabular}{lSSSS}
    \toprule
    \multirow{1}{*}{Section} &
      \multicolumn{1}{c}{Type} &
      \multicolumn{1}{c}{Main goal}
      \\
      \midrule
    \ref{synt_cases} & \text{Synthetic} & \text{Reveal the applicability  of the GEFEST framework}\\
    \bottomrule
        \ref{coastal_sec} & \text{Coastal engineering} & \text{Demonstrate the benefits of combination of different estimators}\\
    \bottomrule
    \ref{micro_sec} & \text{Microfluidics} & \text{Compare deep learning and standard samplers}\\
    \bottomrule
        \ref{heat} & \text{Heat Sources} & \text{Illustrate how to perform generative design with only dataset}\\
    \bottomrule
            \ref{well_sec} & \text{Oil field} & \text{Show how to apply the GEFEST only to specific subproblem}\\
    \bottomrule
  \end{tabular}%
  }
\end{table}

For a fair comparison, we took into account the training time of neural networks in cases when it is necessary. All experiments were conducted using Windows 2008 Server with 32 core units and DGX 1 NVIDIA Cluster with Tesla V100.

\subsection{Synthetic cases} \label{synt_cases}

It is necessary to analyse the applicability of the GEFEST to a problem with different properties before getting down to real-world cases. To achieve this purpose we conducted several synthetic experiments in which we varied different properties of optimal solution: number of polygons, number of vertices, domain size. The main purpose of these experiments is to investigate the relationship between complexity of expected optimal solution and efficiency of search. The hypothesis is that there is a linear relationship for the GEFEST, since it implements a generalized approach that is not specific to some sub-classes of the generative design problem.

In this problem statement, we considered the following GEFEST tools:
\begin{itemize}
  \item $\mathbf{Sampler}$: standard approach (GEFEST Standard Sampler)
  \item $\mathbf{Estimator}$: synthetic estimator based on distance between obtained and reference solution.
  \item $\mathbf{Optimizer}$: biologically-inspired method (Genetic Algorithm).
\end{itemize}

% add details

\textbf{Variable number of objects}

\begin{figure*}[htb]
\centering
` \captionsetup[subfigure]{oneside,margin={0.5cm,0cm}}
  \subfloat[Variable number of objects in optimal solution]{%
    \includegraphics[width=.3\textwidth]{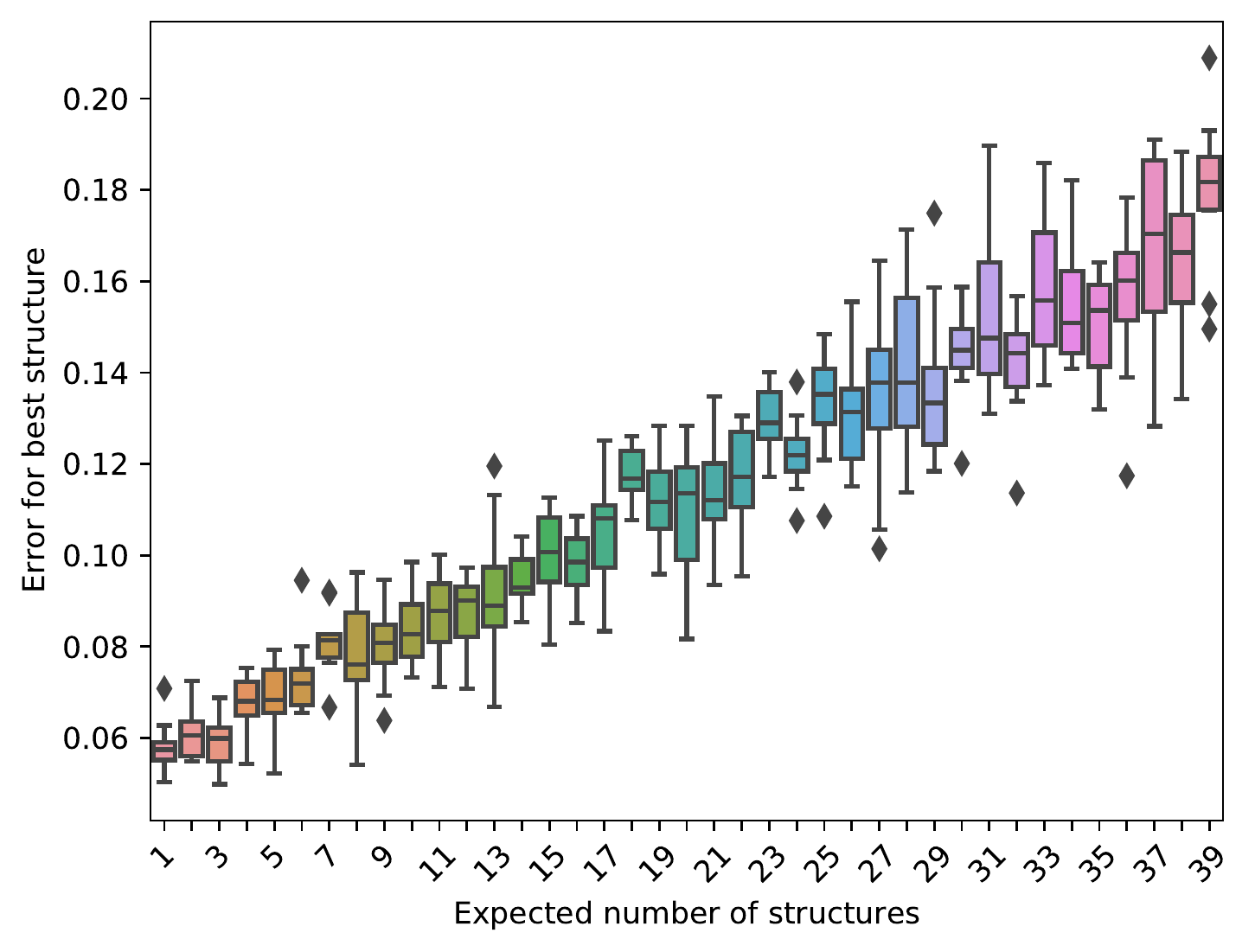}}\hfill
  \subfloat[Variable number of vertices in optimal solution]{%
    \includegraphics[width=.3\textwidth]{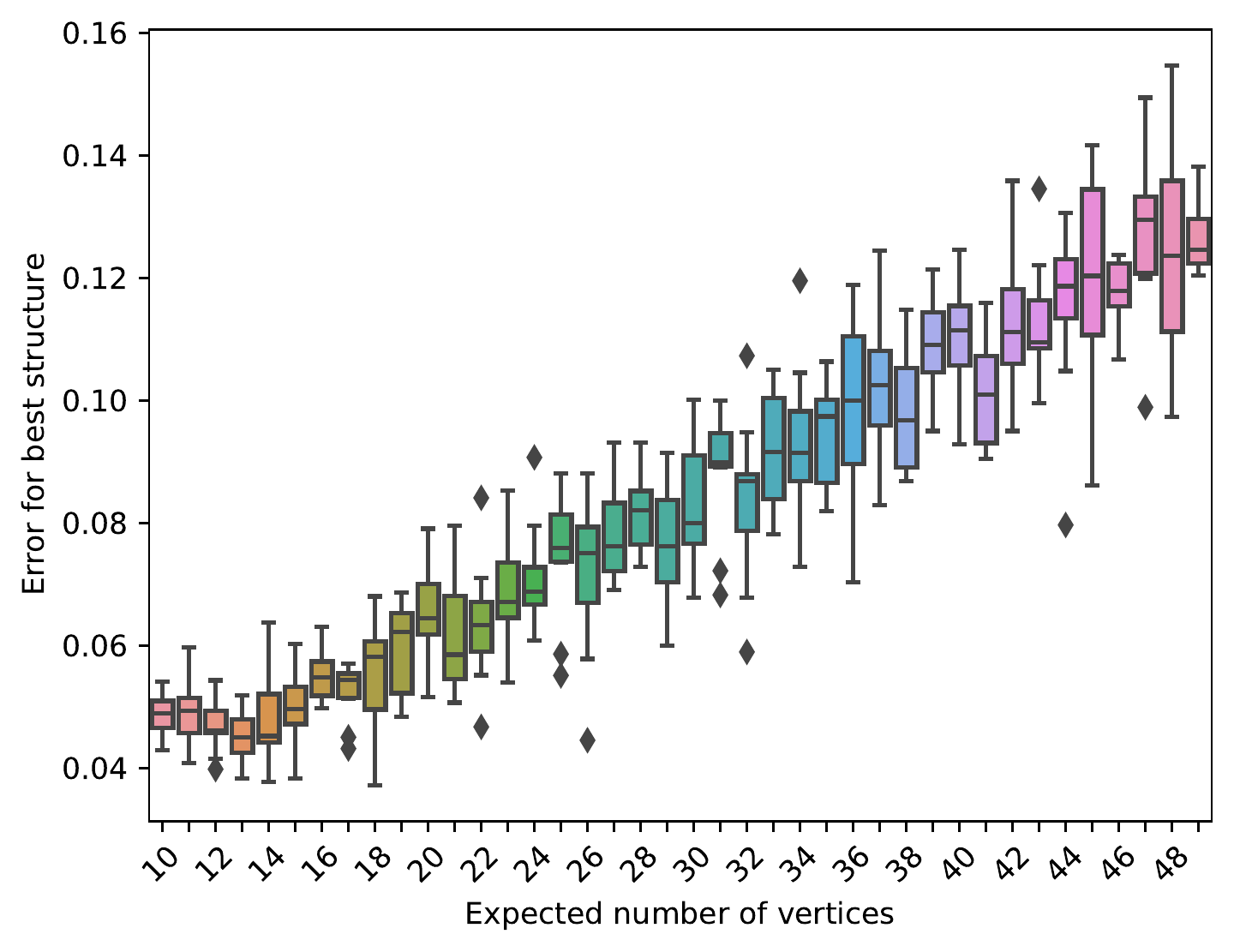}}\hfill
  \captionsetup[subfigure]{oneside,margin={1cm,0cm}}
  \subfloat[Variable size of the search domain]{%
    \includegraphics[width=.3\textwidth]{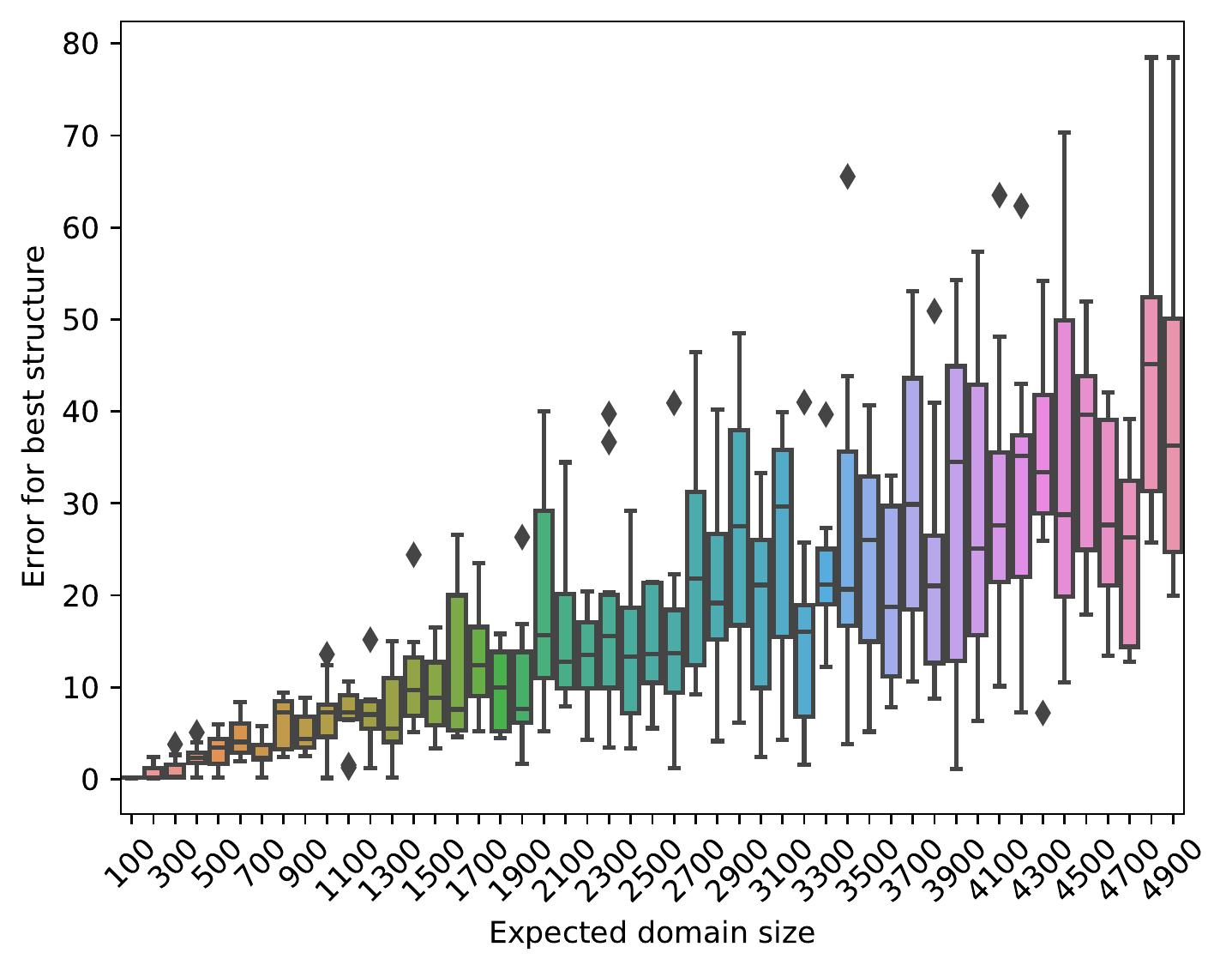}}\hfill
  \caption{The dependence of solution search error from (a) number of objects in optimal solution; (b) number of vertices in optimal solution; (c) size of the search domain}
  \label{fig_synt_exp}
\end{figure*}

The first experiment was devoted to analysis of GEFEST's applicability for reproducing configurations that consists of various number of polygons (from 1 to 30). 
Boxplot that describes effectiveness of the GEFEST for this task is presented in Figure~\ref{fig_synt_exp} (a).

\textbf{Variable number of vertices}

The main difference in the second experiment was the variability of the number of vertices in single polygon (from 10 to 100) instead of number of polygons. Obtained results are presented in Figure~\ref{fig_synt_exp} (b). %add

\textbf{Variable domain size}

Other factor that can affect to effectiveness of the proposed approach is a size of domain that represents the two-dimensional search space. Obtained results are presented in Figure~\ref{fig_synt_exp} (c). %add

As can be seen, the dependence of restoration error to the complexity of configuration can be considered as near-linear for synthetic cases. It empirically confirms the formulated hypothesis. We can conclude that proposed approach is potentially applicable to the various tasks with different properties.

\subsection{Coastal engineering} \label{coastal_sec}
In this subsection we consider real-world design problem from coastal engineering field. This task is dedicated to protecting critical objects (targets) in the water area from natural phenomena (sea waves). Breakwater structures are being developed for these purposes. The main goal is to find a configuration of breakwaters (opened-form polygons in terms of the GEFEST encoding), which minimizes the wave heights at significant points. The cost of breakwaters should also be taken into account. More details about this problem can be found in the existing works \citep{xu2021ecological, nikitin2020multi}

\begin{figure}[h!]
    \centering
    \includegraphics[width=9cm]{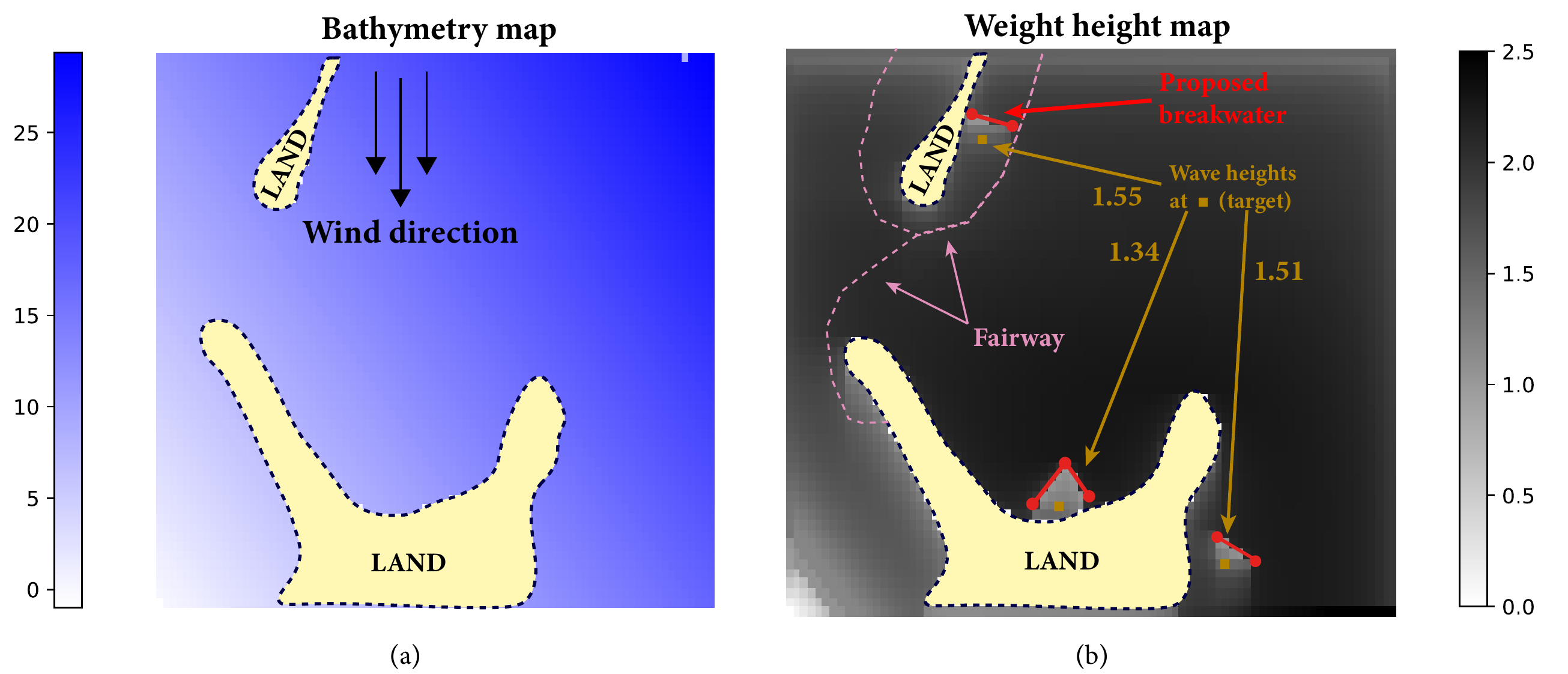}
    \caption{The problem statement for generative design of coastal breakwaters. Bathymetry increases from the lower left corner (0 meters depth) to the upper right (25 meters depth); wave height depends on the position of the breakwaters, maximum value is 2.5. In the right figure baseline solution is presented.}
    \label{bath_h}
\end{figure}

In Figure \ref{bath_h} configuration of our breakwaters design problem is shown. 
 First of all, we specified bathymetry (water depth at each point of the water area), as well as the direction and speed of the wind. In this particular case, we set two land areas, fairways and three targets. These objects are fixed, whereas position of breakwaters (red polygon in Figure \ref{bath_h}) should be optimized to ensure maximum protection of critical objects from sea waves. In addition, crucial constraint is imposed: protecting constructions should not cross fixed objects. Breakwater length, meanwhile, is of great importance, since it directly affects the cost. Therefore, the target variable is two-dimensional $\mathbf{Y} \in \mathbb{R}^2$: the first component is responsible for the sum of wave heights, the second one - for the cost of breakwaters. And the third step of the \textit{generative design procedure} is a multi-objective optimization problem with constraints (see ~\ref{multi-obj-basics} for formal basics). Worth noting that in this case we consider optimization problem (\ref{optim}) in terms of minimization. As baseline solution we choose configuration of breakwaters specially created by ourselves. It is shown in Figure \ref{bath_h} (b). Also, we determined a low value of breakwater protection coefficient. It means that wave height at point will increase quite quickly when moving away from breakwater. 

In this problem statement, we considered the following GEFEST tools:
\begin{itemize}
  \item $\mathbf{Sampler}$: standard approach (GEFEST Standard Sampler)
  \item $\mathbf{Estimator}$: physics-based simulator (Simulating WAves Nearshore) and deep learning (Convolutional Neural Network), 
  \item $\mathbf{Optimizer}$: biologically-inspired methods (differential evolution (DE) and SPEA2, details about the last algorithm are provided in ~\ref{spea2}).
\end{itemize}
The main purpose of this example is to demonstrate how physics-based and deep learning estimators can be combined to enhance the obtained solution. Moreover, we aim to show the opportunity of utilization different optimizers. Thus, we constructed following toolkits based on mentioned GEFEST tools: 1) GSS + SWAN + DE; 2) GSS + SWAN + SPEA2; 3) GSS + CNN/SWAN + SPEA2, and compared them. All toolkits used Algorithm \ref{gen_des} in \textit{traditional manner}, except for the third one, in which we added \textit{extra sampling} to increase the exploration rate. Algorithm \ref{gen_des} was repeated three times for each toolkit with a time limit of 10 hours for one run. Also, we set the population size to 30 and the archive size to 15.

The third toolkit needs to be discussed in more detail. The most time-consuming procedure among considered GEFEST tools is the physics-based SWAN model estimation of the wave heights. In order to reduce the number of SWAN calls we included an additional deep learning estimator, which is noticeably more lightweight. On the other side, deep learning estimator is less accurate. Nevertheless, high accuracy is required only for solutions close to the minimum, whereas in other cases rough approximation is sufficient. More formally, such combination is described in Algorithm \ref{surrogate}. 
\begin{algorithm}[h!]
\caption{Combination of physics based and deep learning estimators }\label{surrogate}
\begin{algorithmic}[1]
\Require $DL, PB, sample$ \Comment{Deep learning, physics-based estimators and sample for estimation}
\Ensure $pf$ \Comment{Performance of sample}
\State $pf = DL.estimate(sample)$ \Comment{Calculating performance of sample using deep learning model}
\If{$pf < threshold$ 
}
\State $pf = PB.estimate(sample)$ \Comment{Recalculating performance using physics-based model if sample is close to the minimum}
\EndIf
\State \Return $pf$
\end{algorithmic}
\end{algorithm}
This procedure allows to execute more optimization steps owing to fewer calls to the physics-based model. The parameter \textit{threshold} provides an opportunity to skip unsuccessful samples, in our case it is equal to 6.0. 

 The deep learning estimator chosen for this problem is based on the generalized architecture (Figure \ref{estimator_arch}). To collect data for its training, Algorithm \ref{gen_des} with the second toolkit (GSS + SWAN + SPEA2) was in progress for 3 hours. As a result, about 700 labeled examples was obtained. After deep learning estimator were prepared, about 6.7 hours left for generative design with the third toolkit. Details about architecture and training procedure see in ~\ref{surr_appendix}.

\begin{figure*}[h!]
    \centering
    \includegraphics[width=17cm, height=5cm]{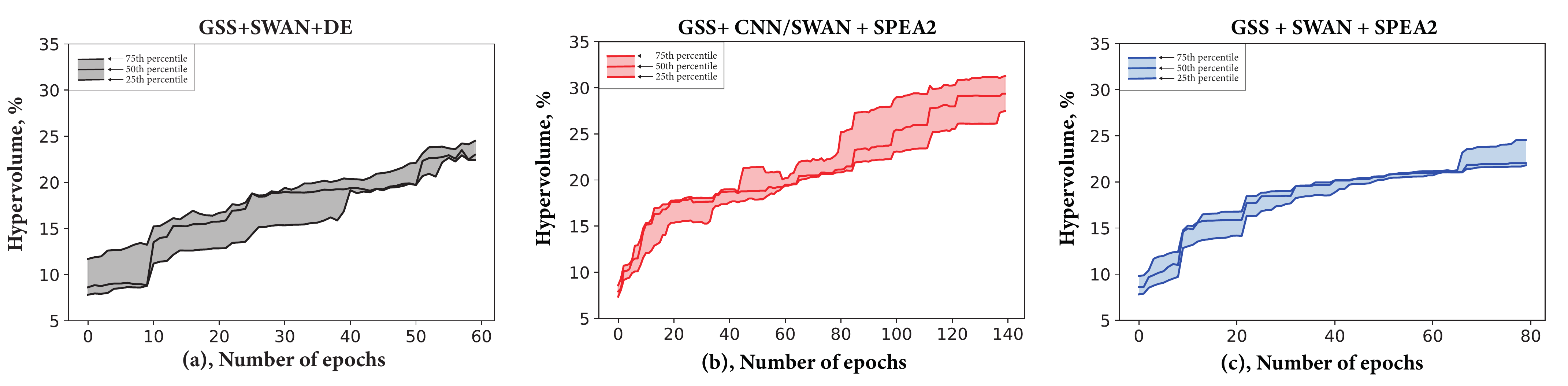}
    \caption{Hypervolume of the population at each step (epoch) of the generative design. The hypervolume was calculated based on three runs, and after 25th, 50th, 75th percentiles were calculated.}
    \label{BW_hv}
\end{figure*}

Results of the experiments is shown in Table \ref{tab:table-name} and Figure \ref{BW_hv}.
\begin{table}[h!]
\caption{\label{tab:table-name} Comparison between different toolkits. The hypervolume is calculated relative to the maximum possible. Here we present the hypervolume at the final step of the generative design. Wave heights are sum of wave heights at all target. In the table arrow $\uparrow$ reflects the larger the better rule, for $\downarrow$ opposite is true.}
\resizebox{\columnwidth}{!}{%
  \begin{tabular}{lSSSS}
    \toprule
    \multirow{2}{*}{Toolkit} &
      \multicolumn{3}{c}{Hyperolume $\uparrow$ (\%), percentile} &
      \multicolumn{1}{c}{Wave heights $\downarrow$ (m)}
      \\
      & {25th} & {50th} & {75th} \\
      \midrule
    GSS+SWAN+DE &22.42 & 22.99 & 24.49 & 4.56\\
    GSS+SWAN/CNN+SPEA2 & $\mathbf{27.47}$ & $\mathbf{29.36}$ &$\mathbf{31.28}$ & \ \ $\mathbf{4.05}$ \\
    GSS+SWAN+SPEA2 & 21.80 & 22.04 & 24.53 & 4.59\\
    Baseline & \textbf{-} & \textbf{-} & \textbf{-} & 4.41\\
    \bottomrule
  \end{tabular}%
  }
\end{table}
It was decided to use hypervolume as the main metric for comparison. For each toolkit we calculated 25th, 50th and 75th percentiles of the latter based on three runs. Moreover, in Table \ref{tab:table-name} the minimum wave heights among three runs is shown. As can be clearly seen from the results, toolkit with deep learning estimator shows the superior performance. This outcome can be explained by the greater number of generative design steps (140 against 60 and 80 for non deep learning estimator toolkits). The integration of the deep learning estimator enables to apply the physics-based model only for important breakwaters and allow more time for further optimization steps. This fact allows to include \textit{extra sampling}. Furthermore, one out of three our approaches surpass the baseline solution.

\begin{figure*}[h!]
    \centering
    \includegraphics[width=17cm]{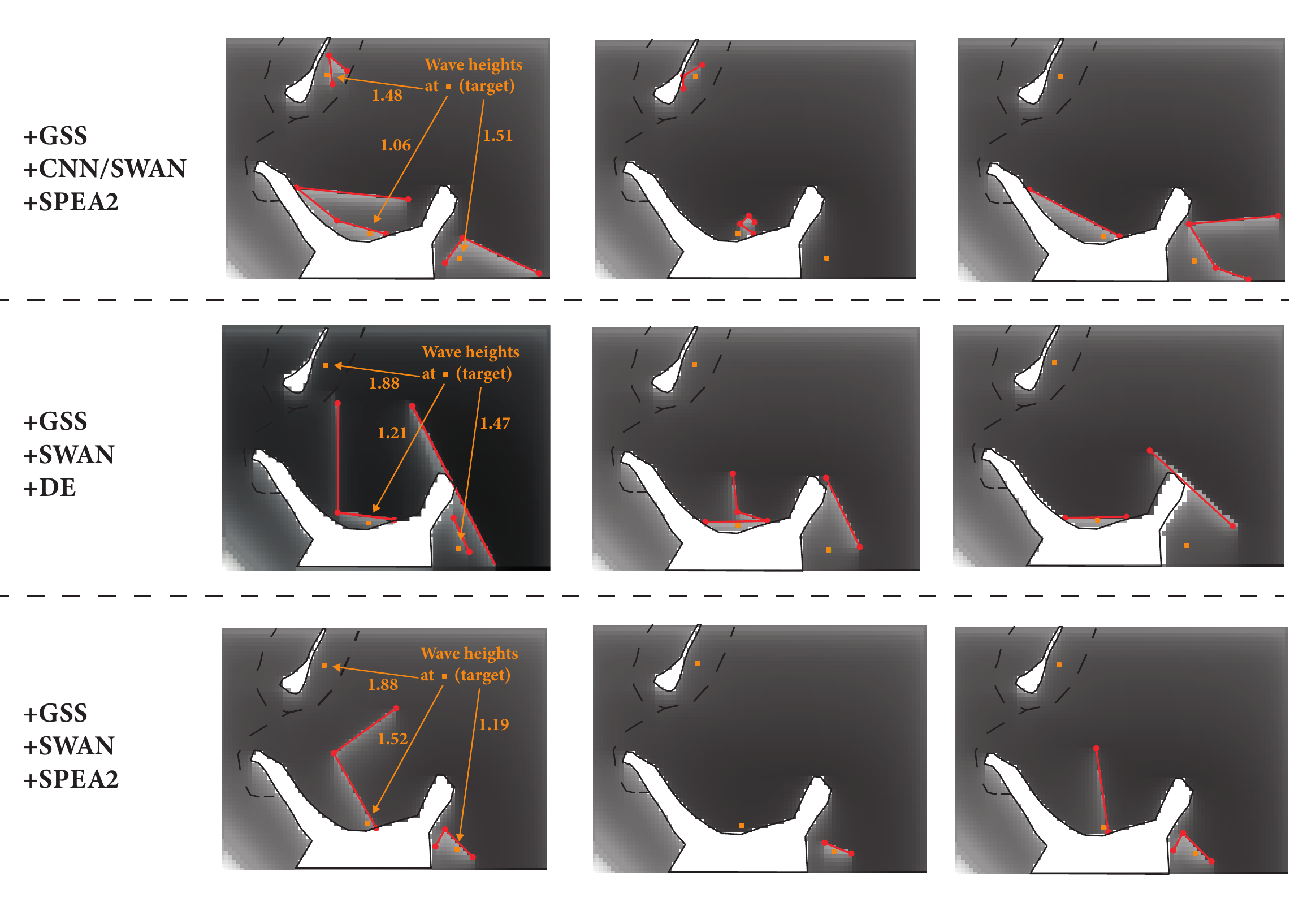}
    \caption{Visualization of some created samples for the breakwaters design problem. Three samples for each toolkit is demonstrated. Left figures in each row shows the best found samples with corresponding wave heights for each target.}
    \label{bw_den}
\end{figure*}

In Figure \ref{bw_den} some created samples for each toolkit are presented.
As can be observed, the only approach permitting to generate the breakwater close to small land includes the deep learning estimator. Furthermore, it excels in sample diversity. Such outcomes is related to the higher exploration rate of the third toolkit compared to others. As was mentioned above, it was achieved by the use of \textit{extra sampling}.

\subsection{Microfluidics} \label{micro_sec}

Microfluidic device is a system with size of about hundreds of micrometers, permeated with several microchannels, the fluid flow through which is investigated. One of the most prominent applications of microfluidics is studying the behaviors of single red blood cells for further biological analysis, disease diagnostics and etc. Conventionally, only certain particles need to be analyzed and thus they should be separated from the unwanted flow components. For these purposes hydrodynamic traps are used. The faster the flow passes between these traps, the higher the trapping probability becomes. For more details, refer to the works \citep{grigorev2022single, man2020microfluidic}.

\begin{figure}[h!]
    \centering
    \includegraphics[width=9cm]{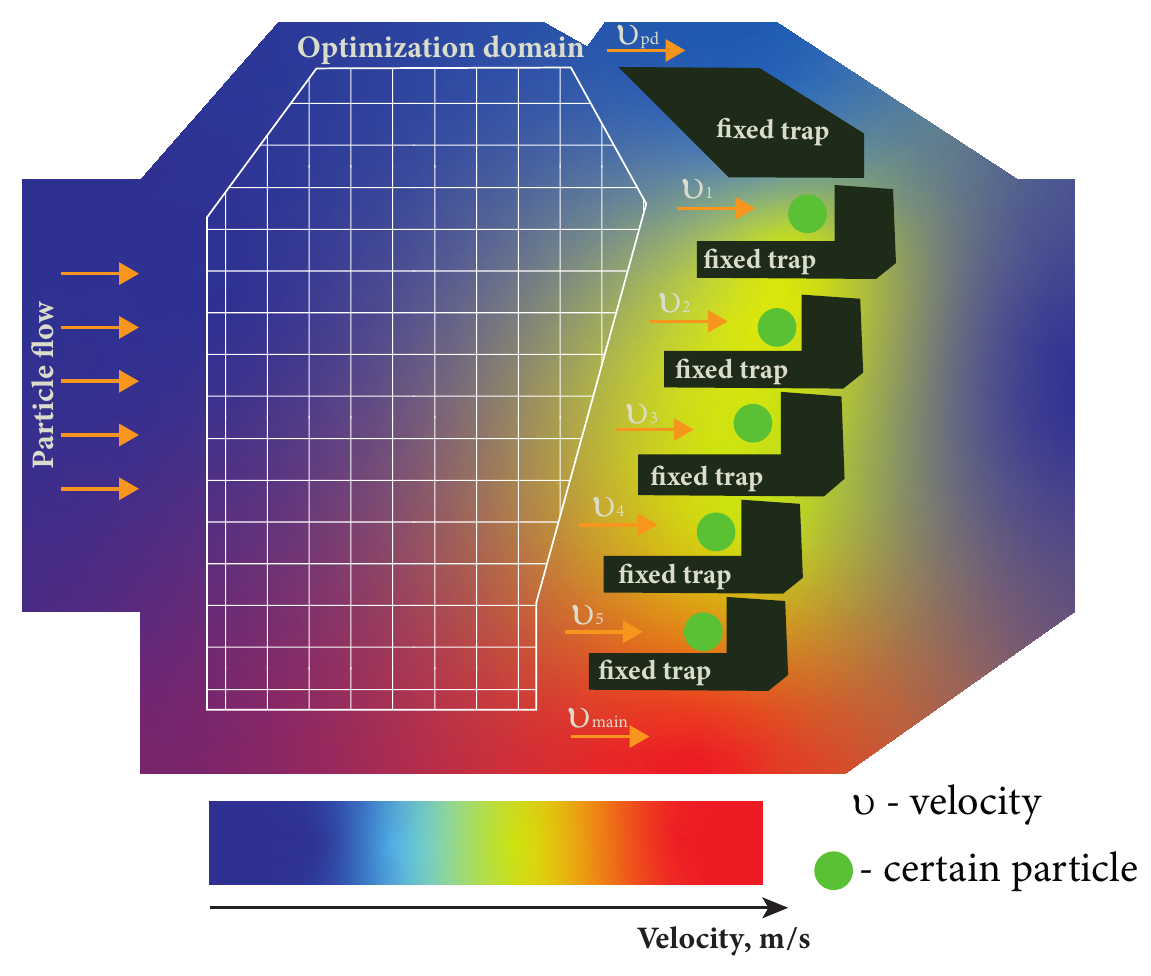}
    \caption{The problem statement for generative design of microfluidic devices. The flow of particles passes through this device. Barriers (closed-form polygon in GEFEST term) can be located inside the optimization domain.}
    \label{micro}
\end{figure}

The general problem statement is shown in Figure \ref{micro}.
In a nutshell, this task consists in barrier construction, which in terms of the GEFEST encoding corresponds to the closed-form polygons. The main purpose is to create polygons inside the optimization domain in such a way that the velocity of particles through fixed traps (1-5) becomes maximal. The increase of the flow rate enhances the capture probability of the certain particle by fixed traps. In addition, the reduction of the velocity through the main and pressure dropping (PD) channels facilitates in achieving the primary goal. Hence, the target variable can be written as follows:
\begin{equation}
    \begin{gathered}
        \mathbf{Y} = \frac{\sum_{i=1}^{5}v_{i}}{v_\text{main} + v_\text{pd}}
    \end{gathered}
\end{equation}
In this case, the optimization problem includes only boundary restrictions without constraints caused by fixed objects within the domain (as it was in the previous section). Note that we use solution from the paper \citep{grigorev2022single} as a baseline.

For the generative design of the hydrodynamic cell traps, we utilized the following GEFEST tools:
\begin{itemize}
  \item $\mathbf{Sampler}$: standard approach (GEFEST Standard Sampler) and deep learning (Generative Neural Network), 
  \item $\mathbf{Estimator}$: physics-based simulator (COMSOL Multiphysics),
  \item $\mathbf{Optimizer}$: biologically-inspired method (Genetic Algorithm).
\end{itemize}
For the closed-form polygon encoding, the application of the deep generative model is more reasonable than in the case of the opened-form polygons. This fact is associated with a greater variability of closed structures. 

The main goal of this study is to reveal the benefits of the deep learning sampler compared to the standard sampler. To this end, we built the following toolkits: 1) GSS + COMSOL + GA; 2) GNN + COMSOL + GA. In order to show the influence of samplers on the generative design results, Algorithm \ref{gen_des} was used in \textit{extra sampling} manner. As in the previous section, the calculation was repeated three times with a time limit of 10 hours for each. The population size was set to 40. 

For the preparation of the deep learning sampler, we collected $100 \ 000$ training objects using standard GEFEST sampler, which took nearly 1.5 hours. In addition, the training of the deep generative model required about 30 minutes. Moreover, beyond the main experiment, we compared different generative neural networks with regard to sample diversity and quality. Details are presented in Section \ref{deep_appendix}. Based on the comparison, for this problem we chose the Adversarial Auto Encoder as deep learning sampler.

\begin{figure}[h!]
    \centering
    \includegraphics[width=7cm]{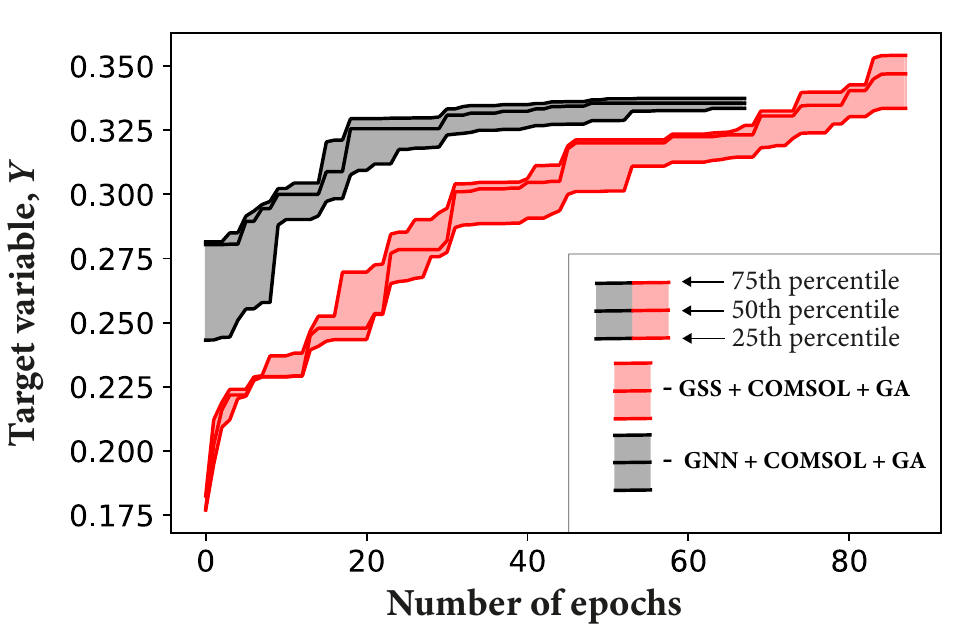}
    \caption{Dependence of target variable on the number of the generative design step for two considered toolkits. The generative design with the normal distribution based toolkit takes greater number of epochs due to the training time of the deep learning sampler which we took into account.}
    \label{micro_res}
\end{figure}

\begin{figure*}[h!]
    \centering
    \includegraphics[width=14cm]{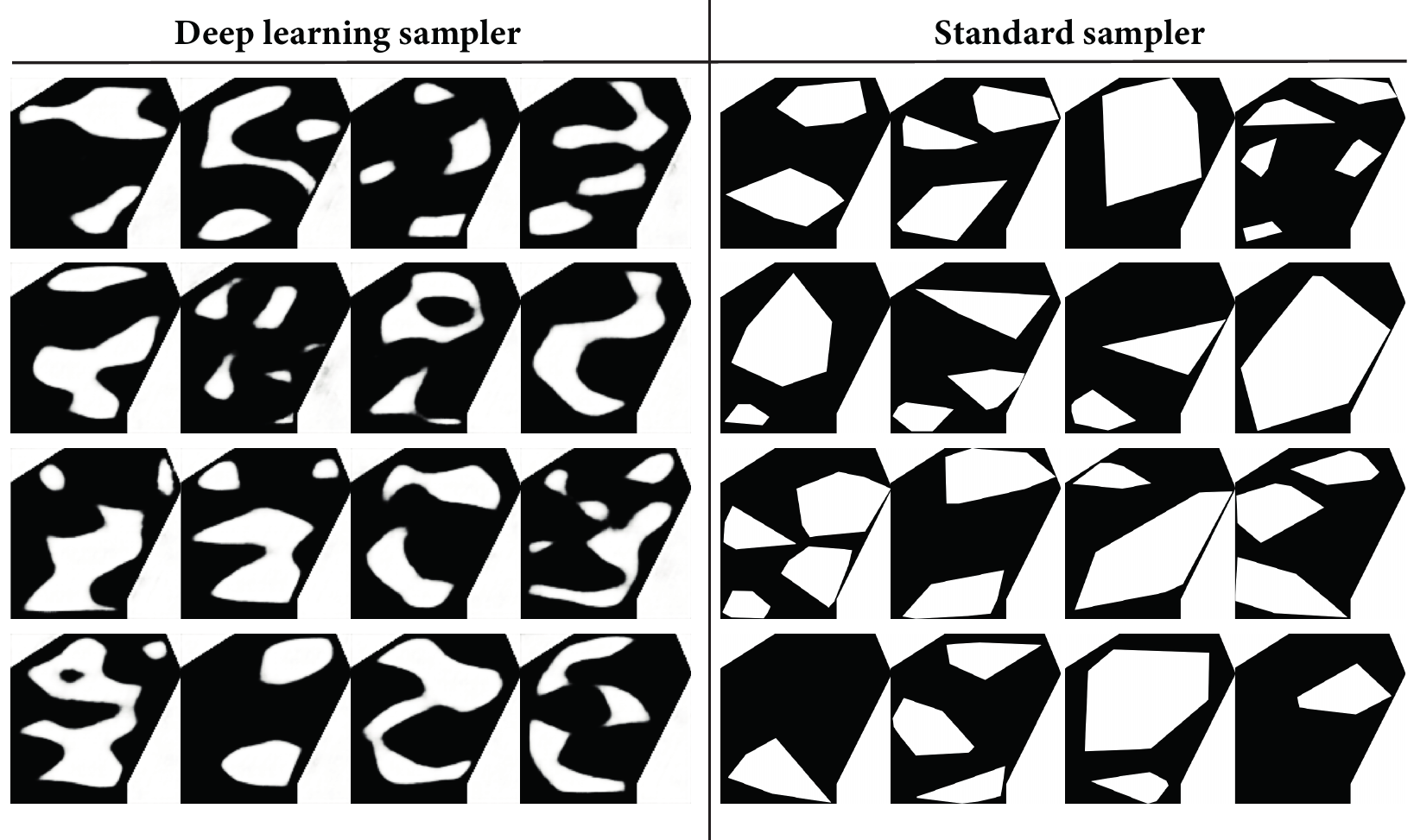}
    \caption{Several objects created using deep learning and standard sampler for microfluidic generative design problem. An adversarial autoencoder was used as a deep learning sampler.}
    \label{micr_samples}
\end{figure*}

The results of the experiments are shown in Table \ref{tab:micr_toolkits} and Figure \ref{micro_res}.
\begin{table}[h!]
\centering
\caption{\label{tab:micr_toolkits} Comparison between final target variable for considered toolkits. Here we presented best value of the target variable among individuals of population in the last epoch. We ran experiment three times and calculated 25th, 50th and 75th percentiles. In the table arrow $\uparrow$ reflects the larger the better rule.}
\resizebox{\columnwidth}{!}{%
  \begin{tabular}{lSSSS}
    \toprule
    \multirow{2}{*}{Toolkit} &
      \multicolumn{3}{c}{Target variable $\uparrow$} &
      \multicolumn{1}{c}{Best target $\uparrow$}
      \\
      & {25th} & {50th} & {75th} \\
      \midrule
    GSS + COMSOL + GA & \text{0.333} & \textbf{0.347} & \textbf{0.354} & \textbf{0.361}\\
    GNN + COMSOL + GA & \textbf{0.333} & \text{0.336} & \text{0.337} & \text{0.339}\\
    Baseline & \text{-} & \text{-} & \text{-} & \text{0.329}\\
    \bottomrule
  \end{tabular}%
  }
\end{table}
As can be seen from Table \ref{tab:micr_toolkits}, the toolkit based on the GEFEST standard sampler performs slightly better. However, the difference between toolkits is negligible. Moreover, both of our approaches surpasses the baseline solution. Actually, it is necessary to highlight another significant fact. As depicted in Figure~\ref{micro_res}, the deep learning sampler based toolkit enables to create higher target variable samples at the initial steps of the generative design. This suggests that the deep learning sampler produces more beneficial primary objects, that is, allows to get the generative design process started from the better samples. In addition, in Figure~\ref{micr_samples} several samples created by both methods are presented. As can be observed, deep learning based approach generates more diverse and unconventional samples. Creation of such objects using the standard sampler would be cumbersome and lengthy procedure.

Besides, we compared sampling time for both approaches. The results is demonstrated in Table \ref{tab:samplers}. 
\begin{table}[h!]
\centering
\caption{\label{tab:samplers} Sampling time of 50, 500 and 1000 objects for deep learning and standard sampler. Each time measurement was repeated 10 times.}
\resizebox{7cm}{!}{%
  \begin{tabular}{lSSSS}
    \toprule
    \multirow{2}{*}{Sampler} &
      \multicolumn{3}{c}{Time (sec.) for sampling} &
      \\
      & {50} & {500} & {1000} \\
      \midrule
    Deep learning & \textbf{0.56$\pm$0.02} & \textbf{6.2$\pm$0.3} & \textbf{13.2$\pm$0.2}\\
    Standard & \text{2.49$\pm$0.17} & \text{25.3$\pm$0.6} & \text{50.6$\pm$1.1} & \\
    \bottomrule
  \end{tabular}%
  }
\end{table}
It can be clearly seen that the deep learning sampler works approximately four times faster than the standard method. Worth noting that the most time-consuming operation in deep learning sampling is the GEFEST polygon encoding, that is, a transformation from the image to the Cartesian coordinate set.

\begin{figure}[h!]
    \centering
    \includegraphics[width=6cm]{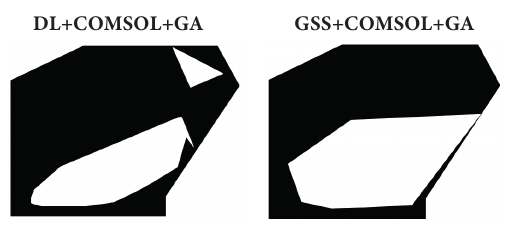}
    \caption{The best objects found by two toolkits for three runs.}
    \label{final_micro}
\end{figure}

In Figure \ref{final_micro} we demonstrated the best objects found using two toolkits. As can be seen, obtained structures closely resemble each other. In both cases the optimization converges to easy-form polygons. However, for the other problems, the opposite may be required.

\subsection{Heat-source systems}
\label{heat}
In this part, we demonstrate the capabilities of the GEFEST framework as a tool for dealing with already prepared datasets. Note that corresponding dataset for the generative design field can be hardly found in open access. Moreover, researchers usually investigate their own specific problems and therefore conventional benchmarks are scarce. Here, we have considered open dataset from related field that is engaged in the heat-source systems investigation \citep{gong2021}. 

Heat-source systems are part of electronic microdevice (micro- or nanometers-sized) that poses a source of heat and therefore temperature field. The control over the temperature distribution within the microdevice (usually called heat management) plays a significant role in practical applications. For instance, real-time knowledge of temperature distribution allows to avoid technical failures, in particular caused by exceeding the critical temperature, and thereby to lengthen the life of the electronic device. However, the distribution across the entire device is commonly unknown. But instead we have the temperature of the monitoring points at our disposal. Thus, the problem of the temperature field reconstruction within the microdevice often attracts the attention of researchers. For more details, refer to the existing works \citep{chen2020heat, gong2021}.

\begin{figure}[h!]
    \centering
    \includegraphics[width=9cm]{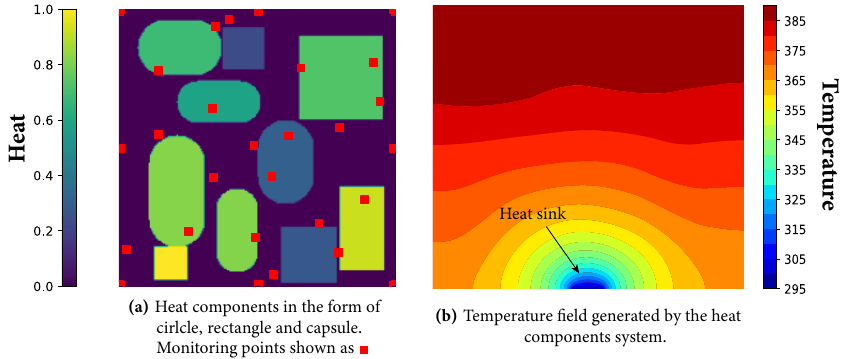}
    \caption{One example from the heat-sources dataset. Heat components (a) and corresponding temperature field (b). }
    \label{teplo_data}
\end{figure}

The chosen dataset consists of $10 \ 000$ examples, each represents two images. We demonstrate one instance in Figure \ref{teplo_data}. 
The left image provides the selected heat-source system inside the electronic microdevice. In this dataset heat-sources are divided into three types according to their shape (circle, rectangle and capsule). Their number remains constant and equals 10, aside from the rare cases when it is reduced by one. Also, each source generates heat evenly, that is, the same value within the component. The right image is a temperature field produced by the given combination of the heat-sources. Note that adiabatic conditions are applied to the boundaries of the device, except for one point, in which heat sink is located. The temperature of the latter is a constant quantity equal to $298$ K. 

In most existing works authors examined the reconstruction of the temperature field using a set of heat-sources. However, we have formulated another problem more specific to the generative design. Our goal lay in the production of heat-source system that insures a minimum average temperature within the device. Thus, the target variable had the following form:
\begin{equation}
    \begin{gathered}
            \mathbf{Y} = \frac{1}{M}\frac{1}{N}\sum_{i=1}^{N}\sum_{j=1}^M{T_{ij},}
    \end{gathered}
\end{equation}
where $M, N$ - number of grid points, $T_{ij}$ - the temperature at a certain point. Worth nothing that in this experiment the optimization problem was considered in terms of minimization. As a baseline solution we have chosen an example from the dataset with minimum target value.

For the solution of the mentioned problem, we selected the following GEFEST tools:
\begin{itemize}
  \item $\mathbf{Sampler}$: deep learning (Generative Neural Network), 
  \item $\mathbf{Estimator}$: deep learning (Convolutional Neural Network),
  \item $\mathbf{Optimizer}$: -.
\end{itemize}
The data-driven methods were chosen by virtue of the fact that in this case we limited ourselves to the dataset only. More precisely, the data generators (that produce new heat-sources) and the physics simulators (that accurately estimate the temperature field for new objects) were left beyond the scope of the described experiment. Thus, the constructed toolkit (GNN + CNN) is completely based on deep learning models. Since in the toolkit the optimizer is absent, Algorithm \ref{gen_des} was ran in \textit{random search} manner. Nevertheless, the deep learning based toolkit can be expanded by including the certain optimizer, but this option will be discussed later.

As in the previous section, the Adversarial Auto Encoder was selected as deep learning sampler. We trained this model on images of the heat components without taking into account its temperature field. The deep learning estimator was learned to approximate the average temperature within the device from the image of heat-sources. Some details are presented in Section \ref{teplo_appendix}. Note that the number of iterations for Algorithm \ref{gen_des} was set to $10 \ 000$.

\begin{figure}[h!]
    \centering
    \includegraphics[width=8cm]{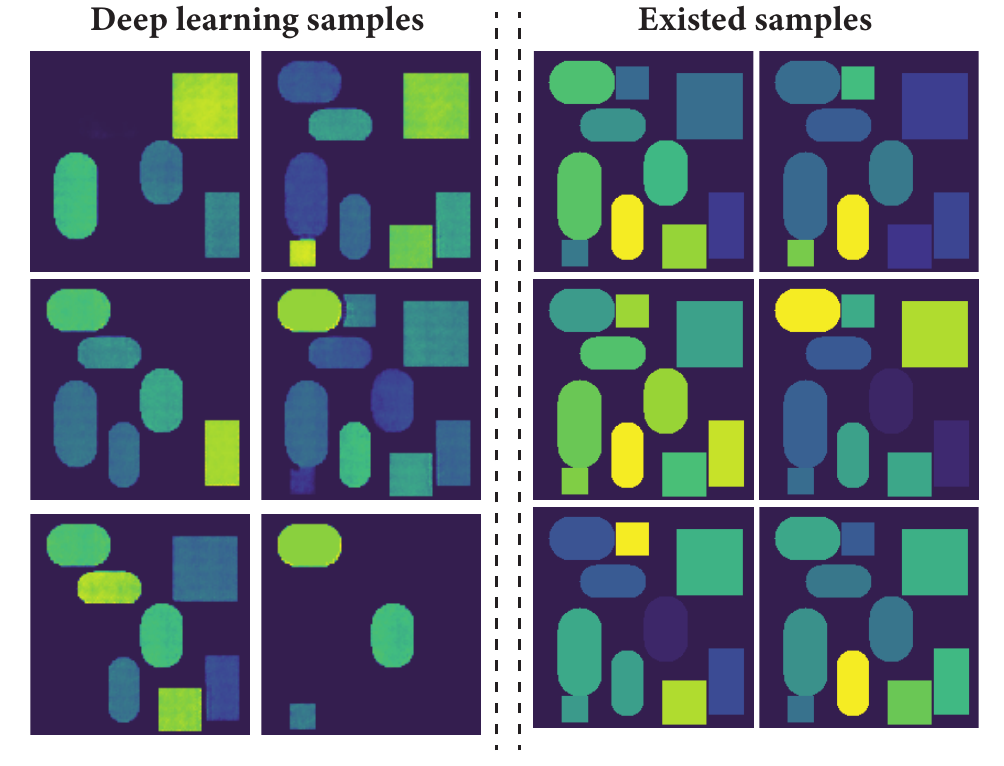}
    \caption{Visualization of several existed samples from the dataset and samples generated by the deep learning model.}
    \label{teplo_generative}
\end{figure}

Before proceeding to the results of the generative design, it is necessary to compare the existed samples and the samples created by the deep learning model, presented in Figure \ref{teplo_generative}. 
As can be observed, the generative model has acquired the ability to generalize. In other words, in addition to existing objects the deep learning sampler generates objects that were not in the original dataset. Consequently, the number of the heat-sources can be vary. Such generalization may result in the production of new unseen samples, which possess a significant value in the generative design.

\begin{figure}[h!]
    \centering
    \includegraphics[width=9cm]{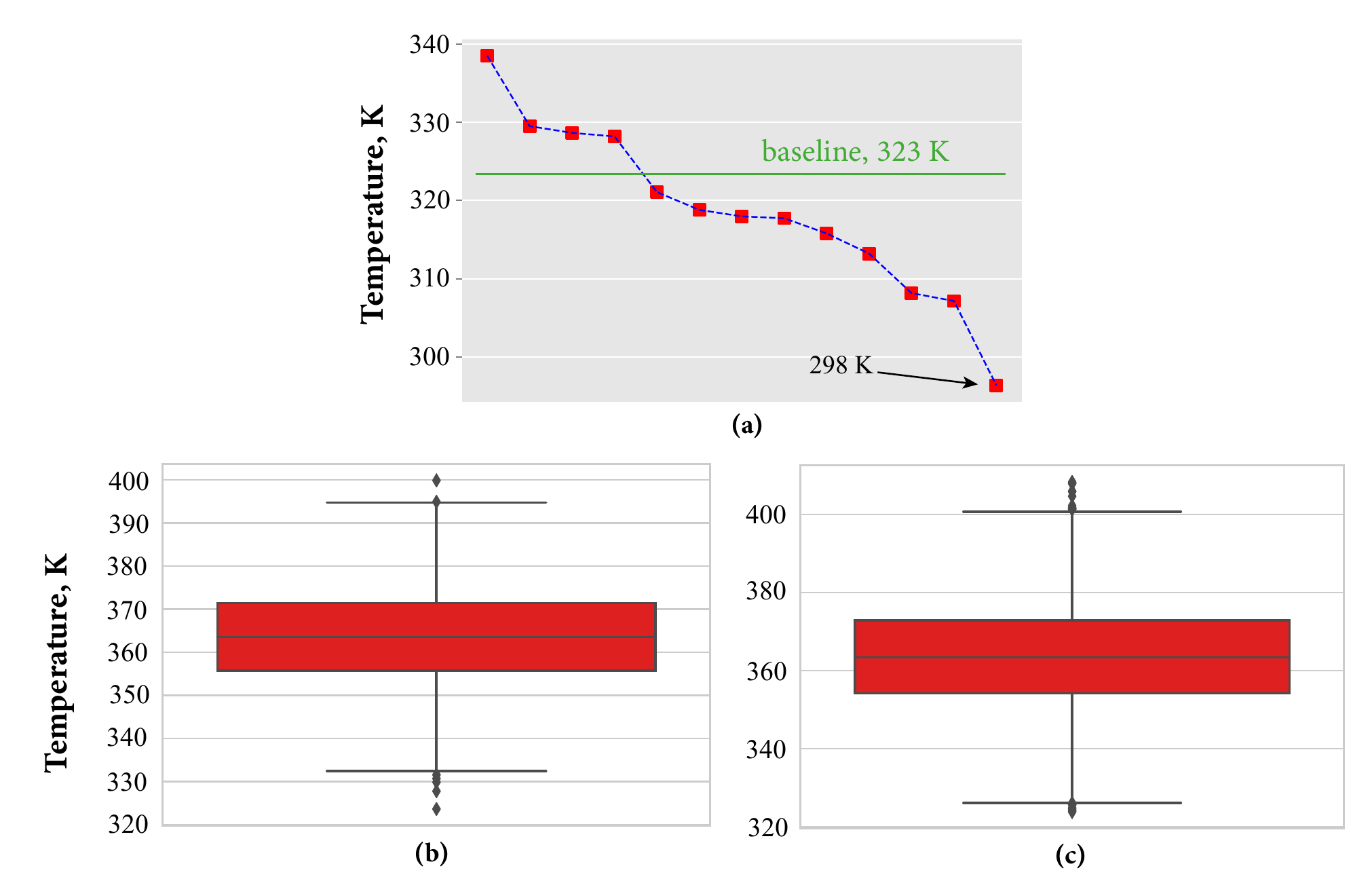}
    \caption{The average temperature of the best samples found during generative design (a). The temperature was calculated by the deep learning estimator. Boxplots constructed based on true average temperature (b) and on predicted temperature (c).}
    \label{teplo_results}
\end{figure}

\begin{figure}[h!]
    \centering
    \includegraphics[width=3cm]{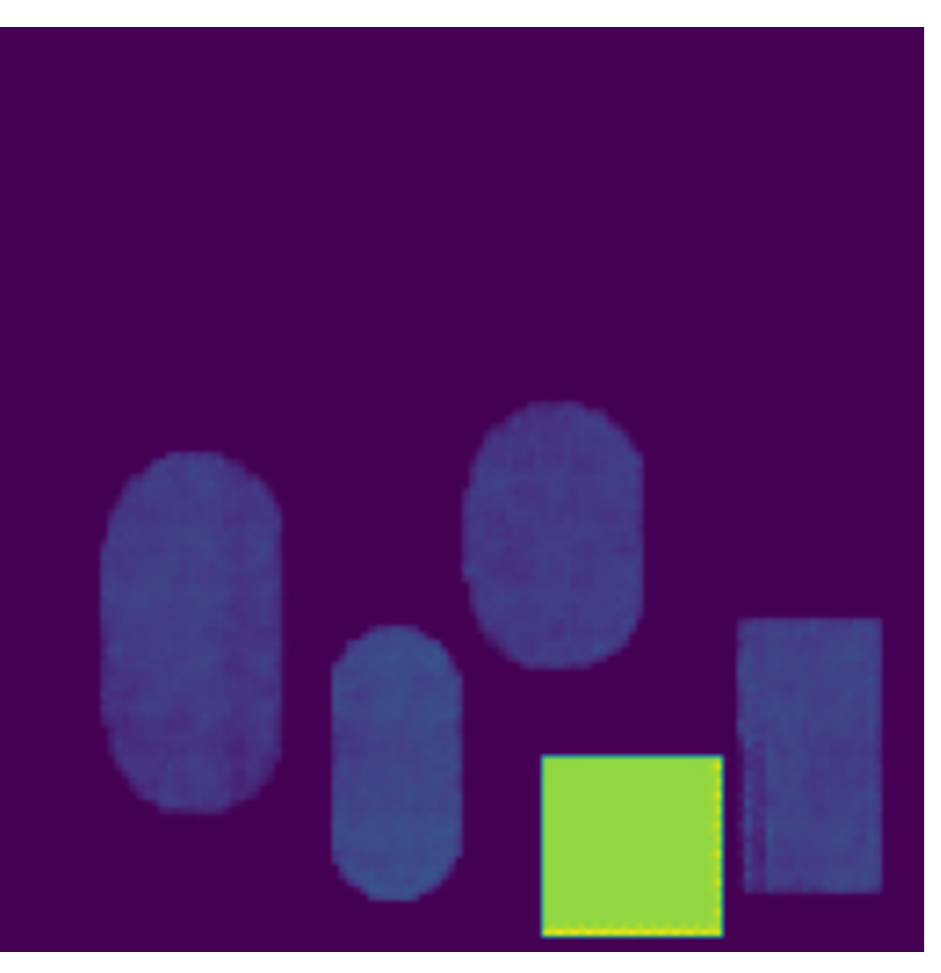}
    \caption{The best found sample providing the minimum average temperature.}
    \label{teplo_final}
\end{figure}

The results of the generative design are presented in Figure \ref{teplo_results}(a) and Figure \ref{teplo_final}.
As can be seen from the comparison between the minimum temperature in the dataset and the minimum value obtained during generative design, we found the configuration of heat-sources reducing the average temperature by 25 degrees. This advantageous configuration is depicted in Figure \ref{teplo_final}. Note that the found object was not present in the original dataset.

\subsection{Oil field planning} 
\label{well_sec}

\begin{figure}[h!]
    \centering
    \includegraphics[width=5cm]{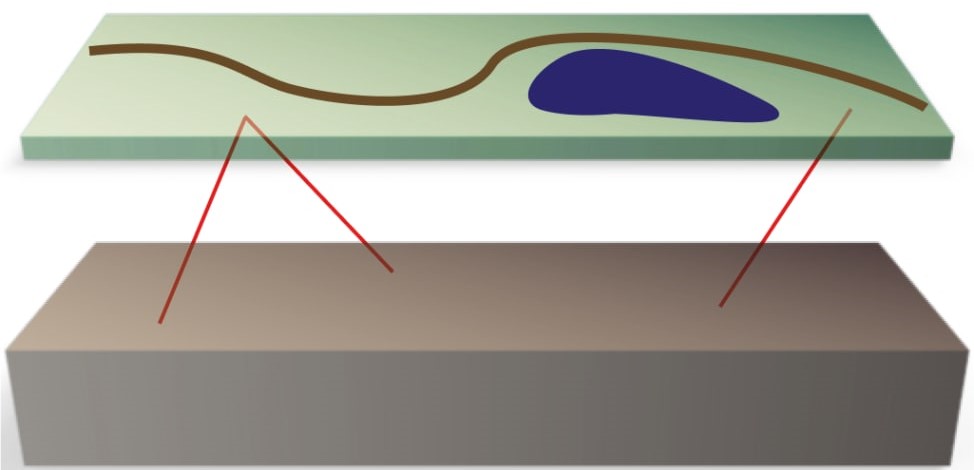}
    \caption{Example of a real field model (wells are highlighted in red, a road is highlighted in brown, and a lake is highlighted in blue).}
    \label{map_field}
\end{figure}

In this part we consider the problem of optimal location of wells and roads in an oil field. Location of wells is the most important stage of field development. So, here we aim to maximize production of field. In modern works \citep{minton2012comparison, tukur2019well, jesmani2020reduced} real limitations in the development
of fields are not considered. More precisely, it does not pay attention to various geographical objects (lakes, swamps or rivers) that make it difficult to build wells and roads. In such a way we consider more general and realistic formulation of the problem. Our optimization task is to find the optimal location of wells and roads taking into account inaccessible areas (wells and roads cannot be inside it) that represent geographical objects. Thus, the joint optimization of wells and roads is being considered, as well as a set of areas that either prohibit the construction of roads and wells, or allow it to be done with a significant penalty. An example of a field with a road, inaccessible area and three wells is shown in Figure~\ref{map_field}.

The main goal of this study is to illustrate how to apply the GEFEST framework only to a subproblem. In other words, the GEFEST should solve one part of the whole task, and another tool should solve the rest. For this purpose, we used a cooperative algorithm that can be divided into two parts:
\begin{itemize}
    \item algorithm for optimizing the location of wells considering roads and inaccessible areas;
    \item an algorithm for optimizing roads considering the location of wells and inaccessible areas.
\end{itemize}
As the first algorithm, various approaches developed on basis of the GA and PSO algorithm were used. At this stage, the basic algorithms have been modified to use the structures of wells and deposits, as well as added consideration of inaccessible points and roads during optimization. As for the second one the GEFEST was used. For both parts of the joint algorithm, special objective functions were used. More precisely, these functions have the following form:
\begin{enumerate}
    \item When optimizing the location of wells, the $NPV$ function is used as the target function, which reflects the economic benefit from the developed field. Optimizing the location of wells \citep{minton2012comparison}:
    \[NPV = \sum\limits_{t=1}^T\frac{CF_t}{(1+r)^t} - C^{capex} - r_{road} \times e_{dist},\]
    where
    \begin{itemize}
        \item $r$ is percentage of profit or discount rate;
        \item $T$ is number of time periods;
        \item $CF_t$ is profit in the period $t$, which is equal to the difference between revenue and expenses in this period. The costs of operating the well are constant, and the profit depends on the volume of oil that was produced in a given period of time in accordance with the simulation of a model;
        \item $C^{capex}$ is cost of well development work. This value takes into account the costs of well construction at the field. The cost depends on the length and gradient of the well;
        \item $r_{road}$ is the coefficient of the cost of one road cell;
        \item $e_{dist}$ is total distance from wells to road.
    \end{itemize}
    Mathematical model was used as a geosimulator that uniformly pumps oil based on only a part of oil in a certain volume. This model was used for calculation $CF_{t}$
    The synthetic deposit $SPE2\footnote{https://www.spe.org/web/csp/datasets/set02.htm\#case2a}$ was used as test data.
    \item A function linearly dependent on the length of the necessary roads is considered as an objective function for road optimization: \[NPV_{road} = r_{road} \times (len_{road} + edist),\]
    where
    \begin{itemize}
        \item $r_{road}$ is the coefficient of the cost of one road cell;
        \item $len_{road}$ is the length of the roads built;
        \item $edist$ is total distance from wells to road.
    \end{itemize}
\end{enumerate}
The cooperative approach of joint optimization of wells and roads at the field consists in the periodic exchange of information between the well optimization algorithm and the road optimization algorithm with the transfer of information about the current optimal locations.

Take a closer look at using the GEFEST framework to solve the problem described in this paragraph. This tool was used to implement the following solutions:
\begin{itemize}
    \item optimizing the location of roads taking into account the location of wells and inaccessible areas;
    \item generation of inaccessible areas.
\end{itemize}
In terms of GEFEST encoding roads are an opened-form polygon with a fixed beginning and end, inaccessible areas are closed-form polygons. For the first solution, we configurated GEFEST toolkit based on following tools:
\begin{itemize}
  \item $\mathbf{Sampler}$: standard approach (GEFEST Standard Sampler),
  \item $\mathbf{Estimator}$: synthetic approach ($NPV_{road}$ function),
  \item $\mathbf{Optimizer}$: biologically-inspired method (Genetic Algo-
rithm).
\end{itemize}
For generation of inaccessible areas we used GEFEST Standard Sampler.

\begin{figure}[h!]
    \centering
    \includegraphics[width=6cm]{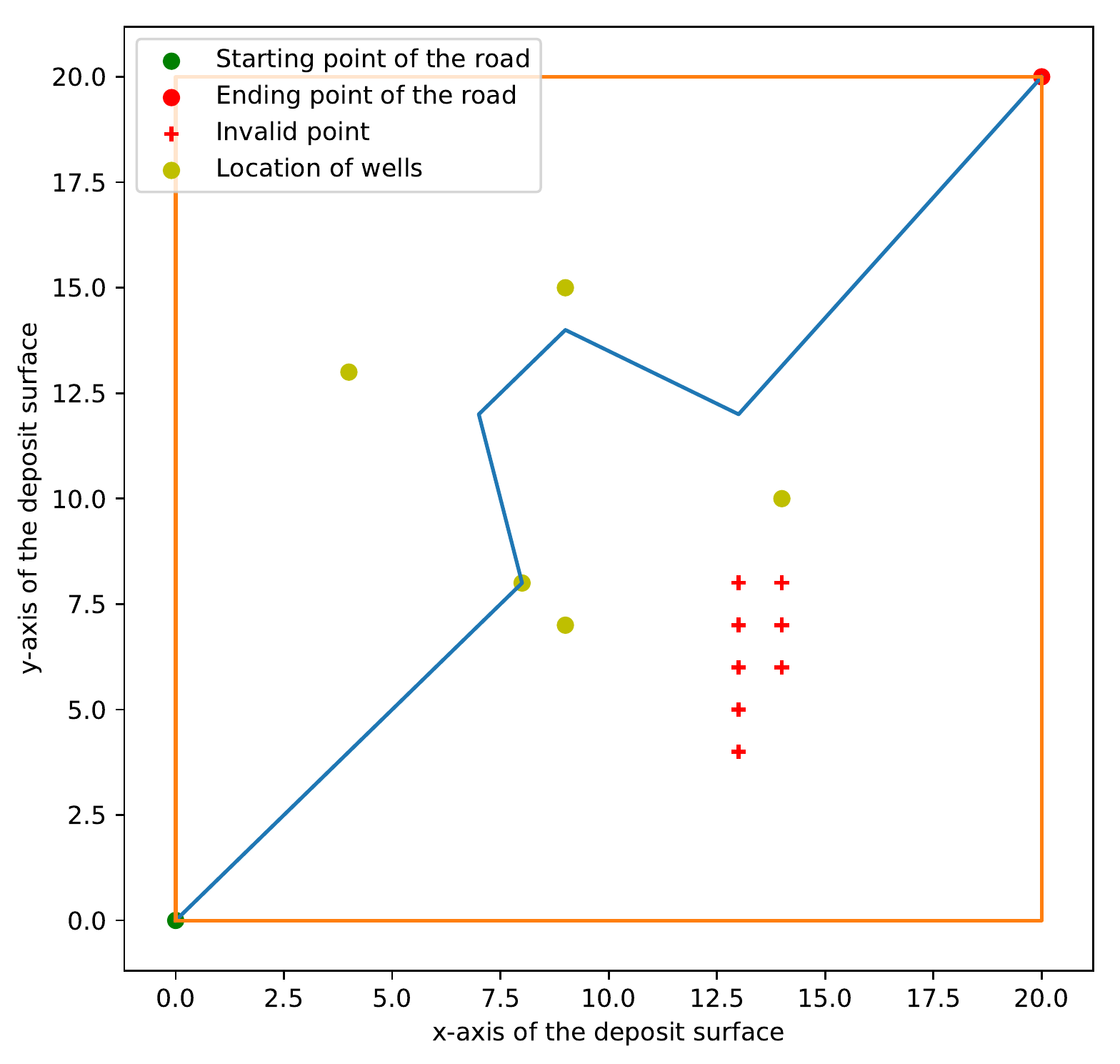}
    \caption{An example of the location of roads, oil wells and inaccessible points on the surface of the field.}
    \label{field_1}
\end{figure}

In Figure~\ref{field_1} an example of the location of objects on the surface of the deposit is shown. In this case, a field with five wells and an inaccessible area of 2\% of the field surface is presented. The blue line shows road, yellow dots represent optimized location of five wells on the surface of the field and red dots represent 8 inaccessible points where it is impossible to build wells and road.

Investigating this task, no developed and published solutions were found. To evaluate the effectiveness of the developed joint algorithm for optimizing wells and roads, taking into account inaccessible areas, a naive approach was developed, which is based on the following steps:
\begin{enumerate}
    \item Optimization of the location of wells without considering roads.
    \item Building a road through optimized well locations without optimizing roads.
\end{enumerate}
This naive approach will be used as a baseline for evaluating the effectiveness of the joint approach.
This algorithm, unlike the cooperative approach, does not imply optimization of the road. To compare baseline and our approach, the following criterion was chosen:
\[
NPV_{joint} = NPV - NPV_{road} - r \times dist, 
\]
where 
\begin{itemize}
    \item $NPV$ is objective function for well placement optimizing algorithm;
    \item $NPV_{road}$ is objective function for road placement optimizing algorithm;
    \item $r$ is road cost coefficient;
    \item $dist$ is total well-road distance.
\end{itemize}
This function allows to take into account economic benefits of developed wells and costs of necessary roads.

In Table \ref{tab:size_areas} and Table \ref{tab:road_cost} results for different size of the inaccessible area and for different road cost coefficient are presented. These tables show the values of the coefficient $K$: 
\[
K = \left(\frac{NPV_{joint}^{cooperative}}{NPV_{joint}^{naive}} - 1\right) * 100\%.
\]
\begin{table}[h!]
\centering
\caption{\label{tab:size_areas} Evaluation of effectiveness of cooperative and naive approaches in dependence of size inaccessible areas.}
\resizebox{9cm}{!}{%
  \begin{tabular}{lSSSSS}
    \toprule
    \multirow{2}{*}{Number of wells} &
      \multicolumn{4}{c}{Size of inaccessible areas} &
      \\
      & {0\%} & {2\%} & {4\%} & {6\%} \\
      \midrule
    3 wells & 4.13\% & 6.41\% & 7.35\%& 10.93\%\\
    5 wells & 4.88\% & 5.16\% & 5.96\%& 8.77\%\\
    7 wells & 4.56\% & 6.05\% & 6.31\%& 9.19\%\\
    \bottomrule
  \end{tabular}%
  }
\end{table}
\begin{table}[h!]
\centering
\caption{\label{tab:road_cost} Evaluation of effectiveness of cooperative and naive approaches in dependence of road cost coefficient.}
\resizebox{9cm}{!}{%
  \begin{tabular}{lSSSSS}
    \toprule
    \multirow{2}{*}{Number of wells} &
      \multicolumn{4}{c}{Road cost coefficient} &
      \\
      & {500} & {1000} & {3000} & {4000} \\
      \midrule
    3 wells & 4.14\% & 6.41\% & 11.95\%& 18.09\%\\
    5 wells & 2.77\% & 4.17\% & 7.19\%& 11.84\%\\
    7 wells & 2.84\% & 4.96\% & 9.67\%& 13.91\%\\
    \bottomrule
  \end{tabular}%
  }
\end{table}
As it can be seen from presented results cooperative approach is constantly more effective than naive approach in both cases. 

\begin{figure}[h!]
    \centering
    \includegraphics[width=8cm]{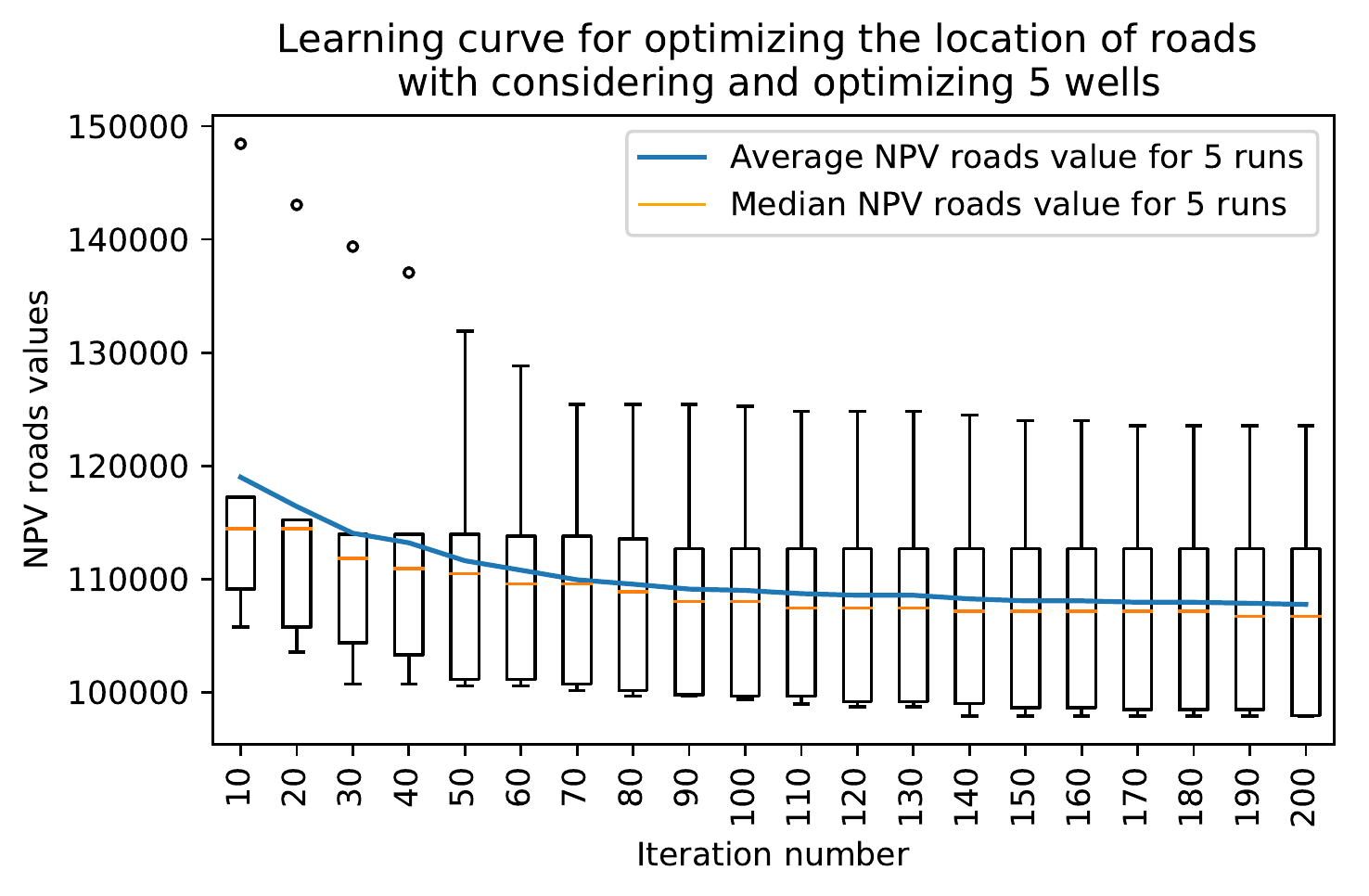}
    \caption{An example of the convergence curve of the objective function for roads when optimizing the location of 5 wells.}
    \label{NPV_iter_plot}
\end{figure}

For convergence of the joint algorithm, 200 iterations of the algorithm were used to optimize the location of roads. An example of an averaged convergence curve is shown in Figure~\ref{NPV_iter_plot}. We presented dependence of $NPV_{road}$ from the iteration number in case of optimization of the location of 5 wells with an area of inaccessible points of size $2\%$.

Thus, the developed cooperative approach with the GEFEST framework proved to be consistently more effective than the naive approach. Also with an increase in the price of roads, the efficiency (K-coefficient) of the cooperative algorithm increases compared to the naive algorithm.

\section{Discussions}
\label{sec_disc}
In the experimental studies, we demonstrated the flexibility of the GEFEST approach, which can be applied to various practical problems. In addition, we revealed the benefits of some tools that can be valuable addressing unexamined real-world problems. Moreover, we showed the opportunity of the GEFEST to generate novel objects in problem limited by dataset.

In the coastal engineering problem, the combination of physics-based model and convolutional neural network in the single approach led to the results surpassed those provided by standard methods. Furthermore, we obtained hard-to-reach extremum, that is, the structure of polygons covering all targets by increasing the exploration rate (\textit{extra sampling} procedure). Such techniques can be easily applied to different  problems taking into account that the deep learning estimator should be sufficiently trained.

In the microfluidic problem, we showed the contribution of the deep learning sampler. The latter allows to create higher performance objects at the initial steps of the generative design. In addition, the generative network can produce not merely regular samples but also diverse and unusual objects in contrast to the standard sampler. Worth noting that these properties depend on the \textit{inductive bias} of the utilized generative neural network. More precisely, it may be the case that a generative model creates samples analogous to those contained in the training set, i.e. reproduce them without any distinguishing features. The choice of a particular model is generally conditioned by the specific problem under consideration. Anyway, despite the appearance of the samples, the inference of the deep learning sampler is faster than for the standard, as was shown earlier. Thus, the deep learning sampler is more preferable, if large set of objects is needed.

In the heat-source systems problem, we demonstrated that the generative design can also be performed using only a prepared dataset. In this case we used Algorithm \ref{gen_des} in \textit{random search} manner, that is, without optimization step. However, another option exists: it is possible to integrate a gradient-based optimizer in the toolkit. In this case the gradient of the deep learning estimator is calculated with respect to the input object. Then, the input updates by gradient descent step. Nevertheless, such procedure can only be applied to neural networks well-trained on huge datasets.

Finally, in the oil field design problem, we illustrated that our framework can be implemented as part of solution to the whole problem. Considering such a situation can be useful for users who want to apply our framework only to a specific subproblem of their task.

\subsection{Limitations}
The primary limitation of our framework is the impossibility of its application for three-dimensional objects that are of the greatest interest in practical fields.

Furthermore, the GEFEST standard sampler is only suitable for production of arbitrary polygonal samples. More precisely, user-defined shapes (circle, rectangle, ellipsis and etc) is infeasible. In such cases it is necessary to utilize other generators or train generative networks on prepared datasets.

Finally, in our approach we considered physical objects neglecting their internal structure. This limitations can be crucial in some generative design problems.

\subsection{Future work}
Future work focuses on extensions of our framework to three-dimensional problems and other types of physical objects. Further, it would be useful to consider dimensionality reduction methods because of an redundant dimension of polygon structure images. Finally, it is essential to explore gradient-based algorithms as a part of generative design concept.

\section{Conclusions}
\label{conclusions}
In this paper, we propose novel open-source framework for generative design of two-dimensional physical objects. The developed approach is based on three general principles: sampling, estimation and optimization. These elements constitute the core of solution to every generative design problem that can be applied to various real-world tasks. We demonstrated the relevance and flexibility of our approach by addressing different applied tasks from ocean engineering, microfluidics, heat-source systems and oil field planning. Furthermore, it was shown that the modification of the general approach ensures the superior performance over baselines. Finally, we revealed the benefits of the tools that, as we believe, can provide objects with refined performance in other generative design problems.

\section{Code and data availability}
\label{sec_code}

The software implementation of all described methods and algorithms as a parts of GEFEST framework is available in the open repository \url{https://github.com/ITMO-NSS-team/GEFEST}. The code and data for experimental studies are available in \url{https://github.com/ITMO-NSS-team/GEFEST-paper-experiments}.

\section{Acknowledgments}

This research is financially supported by The Russian Scientific Foundation, Agreement \#22-71-00094.

\appendix
\section{Multi-objective optimization problem}
Here basics concepts of multi-objective optimization problem with constraints is discussed. Moreover, some details about SPEA2 algorithm are presented.
\subsection{Basics concepts}\label{multi-obj-basics}
We consider a multi-objective optimization problem with constraints, which can be formulated as follows:
\begin{equation}
\begin{gathered}
    \argmin_{x \in X} \mathbf{F}(x) - ? \\
    \text{s.t. } \mathbf{g}(x) = 0 \\
    \ \ \ \ \ \ \mathbf{s}(x) \geq 1
\end{gathered}
\end{equation}
where $\mathbf{F} : X \rightarrow \mathbb{R}^{m}, m \geq 2$ - multi-criteria function; $\mathbf{g}(x) = 0, \mathbf{g}(x) \geq 1$ - constraints that are required to be satisfied. Usually, there is no solution that minimizes all criteria of $\mathbf{F}$ simultaneously. In such cases, the Pareto front is considered. This is a set of all Pareto efficient points in the functional space, more formally \citep{zitzler2001spea2}:
\begin{definition}[Pareto front]
Let $\mathbf{F} : \mathbb{R}^m\rightarrow\mathbb{R}^n$ is a vector function with set of values $\mathbb{Y} = \{y \in  \mathbb{R}^n: y = \mathbf{F}(x), x \in \mathbb{R}^m\}$. The Pareto front is a set $\mathbb{P(Y)} = \{y \in  \mathbb{Y} : \forall y' \neq y \in  \mathbb{Y} \  y \succ y' \}$
\end{definition}
In the definition sign "$\succ$" means Pareto domination:
\begin{definition}[Pareto domination]
Let $y^{1}, y^{2} \in \mathbb{R}^{m}$, $y^{1}$ Pareto dominates $y^{2}$ ($y_{1} \succ y_{2}$) $\iff$ $\forall i = 1 \dots m \ y_{i}^{1} \leq y_{i}^{2}$ and $\exists \ j = 1 \dots m : y_{j}^{1} < y_{j}^{2}$ 
\end{definition}

\begin{figure}[h!]
    \centering
    \includegraphics[width=7cm]{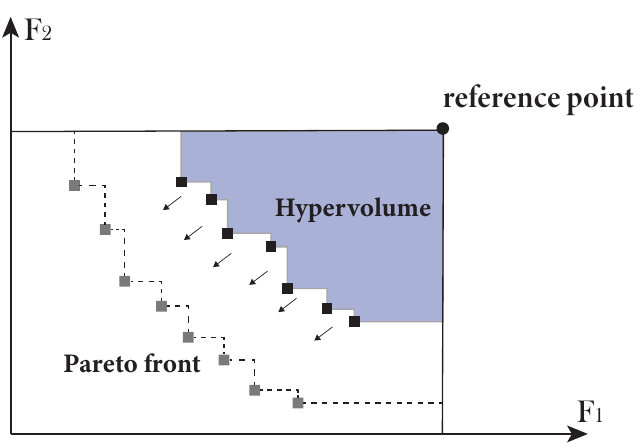}
    \caption{Hypervolume definition. As the algorithm converges, the Pareto front tends to the lower left corner, increasing the hypervolume.}
    \label{Hypervol}
\end{figure}

The main measure of convergence of a multi-objective optimization algorithm is a hypervolume, which can be defined as area between Pareto front and reference point as shown in Figure \ref{Hypervol}.
%все-таки предлагаю картинку убрать - ну или сделать более специфичной именно для ГД.
As the algorithm converges, the Pareto front aspires to left bottom corner (in case minimization problem), thus hypervolume should increase.
\subsection{SPEA2 algorithm} \label{spea2}
The Strength Pareto Evolutionary Algorithm 2 (SPEA2) is an evolutionary based algorithm for approximating of the Pareto front. In the SPEA2 two types of populations are considered: the archive $\mathbf{A}$ and the population $\mathbf{P}$. The archive contains individuals non dominated by any others. In other words, archive is necessary to preserve elitism. The population $\mathbf{P}$ allows to bring new individuals via genetic transformation (mutation, selection, crossover). A core of the SPEA2 is a fitness calculation based on raw and density functions:
\begin{equation}
    \begin{gathered}
        \mathbf{F}(I) = \mathbf{R}(I) + \mathbf{D}(I), 
    \end{gathered}
\end{equation}
where $I$ - individual from population, $\mathbf{R}, \mathbf{D}$ - raw and density functions, which are defined as follows:
\begin{equation} \label{raw}
    \begin{gathered}
        \mathbf{R}(I) = \sum_{I' \in \mathbf{A} \cup \mathbf{P}} \left[I' \succ I \right] \cdot \mathbf{S}(I'), \\
        \mathbf{S}(I') = \text{\#} \ \{I \ | \ I \in \mathbf{A} \cup \mathbf{P}:  I' \succ I\}.
    \end{gathered}
\end{equation}
\begin{equation} \label{density}
    \begin{gathered}
        \mathbf{D}(I) = \frac{1}{d_{I}^{k} + 2}.
    \end{gathered}
\end{equation}
In \ref{raw} ${S}(I')$ is a strength, which defines the number of individuals dominated by $I'$, in \ref{density} $d_{I}^k \ - $ distance from $\mathbf{I}$-th individual to its ${k}$-th neighbour in the functional space. For non dominated solutions $\mathbf{R}(I) = 0$, whereas the density function is necessary to increase a diversity of population

The main loop of the SPEA2 is shown in Algorithm \ref{SPEA2}, \citep{zitzler2004tutorial}.
\begin{algorithm}[h!]
\caption{SPEA2}\label{SPEA2}
\begin{algorithmic}[1]
\Require $M, N, T$ \Comment{Population, archive size and maximum number of steps}
\Ensure $A$ \Comment{Archive population}
\State $Random$ $initialization$
\State $Fitness$ $calculation$ \Comment{Assigning fitness to each individual from $\mathbf{P} \cup \mathbf{A}$}
\State $Environmental$ $selection$ \Comment{Filling the archive with non-dominant solutions}
\State $Termination$ \Comment{If stopping criterion is satisfied then return $\mathbf{A}$}
\State $Mating$ $selection$ \Comment{Perform selection operator on $\mathbf{P} \cup \mathbf{A}$}
\State $Variation$ \Comment{Apply crossover and mutation operators to the selected population}
\end{algorithmic}
\end{algorithm}

\section{Deep learning models}
Here architectures and training process of the deep learning models used in the experimental studies are discussed.

\subsection{Coastal engineering estimator} \label{surr_appendix}
The deep learning estimator takes the image of the breakwaters in an input corresponding to Figure \ref{estimator_arch}. The estimator consists of three convolutional layers with $L2$ regularization, one GlobalMaxPooling layer and two fully-connected layers. The total number of parameters is equal to $372\ 449$. 

\begin{figure}[h!]
    \centering
    \includegraphics[width=9cm]{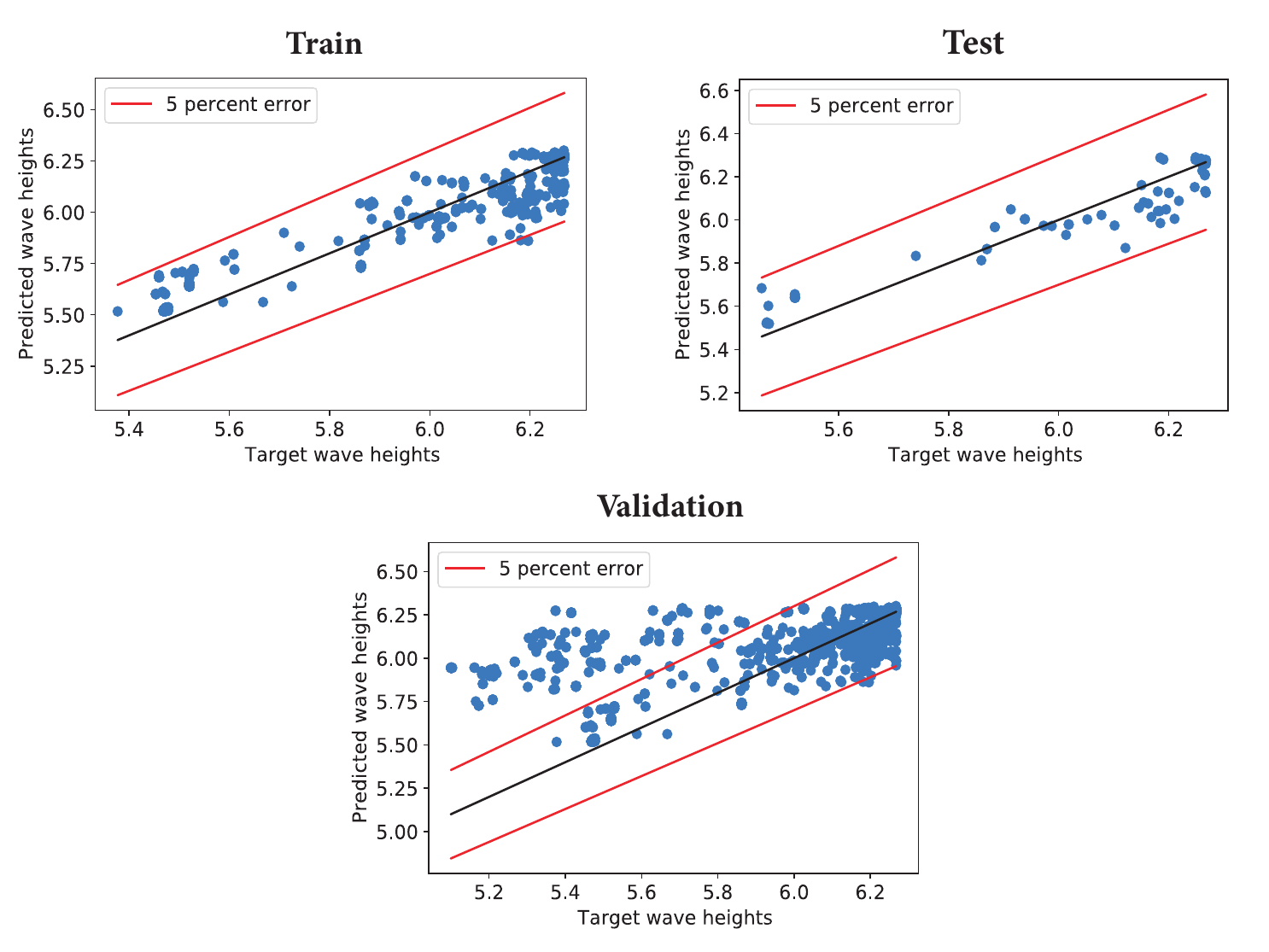}
    \caption{Correlation plots between predicted wave heights and simulated wave heights for training, test and validation samples.}
    \label{corr_plot}
\end{figure}

For evaluation of the convolutional neural network, we have used a validation set created beyond the time of generative design. In Table \ref{surrogate_table} and Figure \ref{corr_plot} the results of the deep learning estimator approximation of wave heights are shown.
\begin{table}[h!]
\centering
\caption{\label{surrogate_table} Losses of the deep learning estimator and sizes for training, testing and validation datasets.}
\resizebox{4cm}{!}{%
  \begin{tabular}{lSSSS}
    \toprule
    \multirow{1}{*}{Dataset} &
      \multicolumn{1}{c}{MAE} &
      \multicolumn{1}{c}{MAPE} &
      \multicolumn{1}{c}{Size}
      \\
      \midrule
    train & \text{0.07} & \text{1.34} & \text{705}\\
    test & \text{0.08} & \text{1.42} & \text{79} \\
    validation & \text{0.20} & \text{3.69} & \text{2376}\\
    \bottomrule
  \end{tabular}%
  }
\end{table}
As can be seen from Figure~\ref{corr_plot}, some predictions of the deep model is prone to overestimated. However, in our problem high accuracy of prediction is not required.

\subsection{Microfluidic deep learning sampler} \label{deep_appendix}
To train the deep learning sampler, we collected $100 \ 000$ examples using standard GEFEST sampler, some of them are shown in Figure \ref{micr_samples}. Produced objects pose right-form polygons without self-intersection, intersection with other structures and out-of-bound parts. The number of polygons within the domain varied from 1 to 7. It is worth noting that GEFEST standard sampler has no restrictions on the number of polygons. However, in case of a large number of the latter such straightforward procedure will be computational expensive. 

On the gathered dataset we trained and compared several deep generative models: Variatonal Auto Encoder \citep{kingma2013auto}, Adversarial Auto Encoder \citep{makhzani2015adversarial}, Variational Normalizing flows \citep{rezende2015variational} and Variational Generative Adversarial Network \citep{larsen2016autoencoding}. In the inference (or sample creation) mode all mentioned deep learning samplers are based on the architecture (Figure \ref{sampler_arch}). For these models we constructed the same backbone: six convolutional layers with batch normalization and ReLU activation with the exception of the last one where tanh function was used. The total number of parameters was generally about 4M, the accurate value depends on the specific model. 

The key criteria in selection of a generative neural network are diversity, quality of samples and speed of inference. The latter turned out to be equal for outlined models due to identical sampling procedure (Figure \ref{sampler_arch}). Diversity and quality of samples can be estimated using Frechet Inception Distance. The results are presented in Table \ref{deep_samplers_app}. 
\begin{table}[h!]
\centering
\caption{\label{deep_samplers_app} Frechet Inception Distance for different deep learning samplers in the microfluidic generative design problem. We used $10 \ 000$ samples to evaluate the FID.}
\resizebox{\columnwidth}{!}{%
  \begin{tabular}{lSSSS}
    \toprule
    \multirow{1}{*}{} &
      \multicolumn{1}{c}{\textbf{Adversarial Auto Encoder}} &
      \multicolumn{1}{c}{Variational Auto Encoder} &
      \multicolumn{1}{c}{Normalizing flows} &
      \multicolumn{1}{c}{Variational GAN}
      \\
      \midrule
    FID & \textbf{277} & \text{308} & \text{305} & \text{407}\\
    \bottomrule
  \end{tabular}%
  }
\end{table}
It is evident that the best performance was shown by the Adversarial Auto Encoder. The AAE produced samples are demonstrated in Figure \ref{bw_den}. Based on calculated FID values, we decided on AAE as a deep learning sampler in the microfluidic problem.

\subsection{Deep learning estimator for heat-source systems} \label{teplo_appendix}
As convolutional backbone in the deep learning estimator we took the EfficientNet model \citep{tan2019efficientnet} pretrained on ImageNet dataset. The total number of parameters was equal to $7 \ 254 \ 843$. We divided the initial dataset into train (80 \%) and test (20 \%). The number of the training epoch was equal to 10. The results of the prediction is presented in Table \ref{heat_table}

\begin{table}[h!]
\centering
\caption{\label{heat_table} Losses of the deep learning estimator and sizes for training and testing.}
\resizebox{4cm}{!}{%
  \begin{tabular}{lSSSS}
    \toprule
    \multirow{1}{*}{Dataset} &
      \multicolumn{1}{c}{MAE} &
      \multicolumn{1}{c}{MSE} &
      \multicolumn{1}{c}{Size}
      \\
      \midrule
    train & \text{4.66} & \text{40.78} & \text{8000}\\
    test & \text{5.03} & \text{49.34} & \text{2000} \\
    \bottomrule
  \end{tabular}%
  }
\end{table}

\bibliography{sample}

\end{document}